\documentclass[11pt]{article}
\usepackage[final]{nips_2017}
\usepackage{amsmath,amsfonts,amsthm,amssymb,graphicx}
\usepackage[table,xcdraw]{xcolor}
\usepackage{booktabs,longtable,float,adjustbox,listings,caption, subcaption} 
\usepackage{natbib}
\usepackage{color,graphics,fancyhdr, lipsum,setspace,soul}
\usepackage[table,xcdraw]{xcolor}
\usepackage{multirow,multicol,verbatim}
\usepackage{enumerate,filecontents}
\usepackage[english]{babel}
\usepackage[normalem]{ulem}

\usepackage[T1]{fontenc}    
\usepackage{hyperref}       
\usepackage{url}            
\usepackage{booktabs}       
\usepackage{amsfonts}       
\usepackage{nicefrac}       
\usepackage{microtype}      
\usepackage[toc,page]{appendix}
\usepackage{comment}
\usepackage{algorithm} 
\usepackage{algpseudocode}

\usepackage{enumitem} 
\usepackage{wrapfig}  

\newcommand{\add}[1]{{\color{black} {#1}}}
\definecolor{dkgreen}{rgb}{0,0.6,0}
\definecolor{gray}{rgb}{0.5,0.5,0.5}
\definecolor{mauve}{rgb}{0.58,0,0.82}
\usepackage{todonotes}

\title{Stock-out Prediction in Multi-echelon Networks}

\author{
	Afshin Oroojlooyjadid 
	\\
	Lehigh University\\
	Bethlehem, PA 18015 \\
	\texttt{oroojlooy@lehigh.edu} \\
	\And
	Lawrence Snyder
	\\
	Lehigh University\\
	Bethlehem, PA 18015 \\
	\texttt{larry.snyder@lehigh.edu} \\
	\And
	Martin Tak\'a\v{c}
	\\
	Lehigh University\\
	Bethlehem, PA 18015 \\
	\texttt{takac.mt@gmail.com} \\
}
\begin{document}

	\maketitle
	
\vspace{1 in}
\begin{abstract}
In multi-echelon inventory systems the performance of a given node is affected by events that occur at many other nodes and in many other time periods. For example, a supply disruption upstream will have an effect
on downstream, customer-facing nodes several periods later as the disruption "cascades" through the system. 
There is very little research on stock-out prediction in single-echelon systems and (to the best of our knowledge) none on multi-echelon systems. 
However, in real the world, it is clear that there is significant interest in techniques for this sort of stock-out prediction. Therefore, our research aims to fill this gap by using deep neural networks (DNN) to predict stock-outs in multi-echelon supply chains.

\end{abstract}

\section{Introduction}\label{sec:prdt_introduction}

\begin{figure}[ht]
	\centering	
	\includegraphics[scale = 0.25]{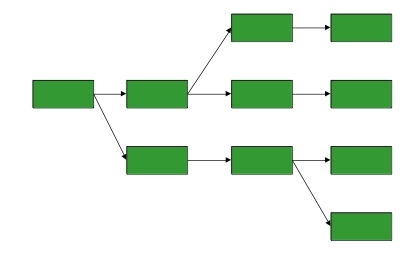}		
	\caption{A multi-echelon network with 10 nodes}
	\label{prdt_sample_multi_echelon}
\end{figure}

A multi-echelon network is a chain of nodes that aims to provide a product or service to its customers. Each network consists of production and assembly lines, warehouses, transportation systems, retail processes, etc., and each of them is connected at least to one other node. 
The most downstream nodes of the network face the customers, which usually present an external stochastic demand. 
The most upstream nodes interact with third-party vendors, which offer an unlimited source of raw materials and goods.
An example of a multi-echelon network is shown in Figure \ref{prdt_sample_multi_echelon}, which depicts a distribution network, e.g, a retail supply chain. 

The supply chain manager's goal is to find a compromise between the profit and service level (i.e. a number between zero and one that determines the percent of the customer's orders that are satisfied on time) to its customers. 
For example, a retail network may decide to change the number of retail stores to increase its service availability and create more sales, which also results in a higher cost for the system. 
In this case, the relevant decisions are how many, where, and when they should be opened/closed to maximize the profit. 
Facility location and network design are the common mathematical programming problems to provide the optimal decision in those questions.
Similarly, the problems in production and inventory systems, are where, when, how, and how much to produce or order of which item. 
Scheduling and capacity management are common problems in this area.
Also, distribution systems must decide when, where, how, and how much of which item should be moved. 
The transportation problem is the most famous problem that answers these questions.
In well-run companies, there are multiple systems that optimize those problems to provide the best possible balance between service level and profit.
In this paper, we focus on inventory management systems to provide an algorithm that answers some of the questions in an environment with stochastic demand.

Balancing between the service level and profit in an inventory system is equivalent to balancing the stock-out level and holding safety stock.
Stock-outs are expensive and common in supply chains. For example, distribution systems face $6\%-10\%$ stock-out for non-promoted items and $18\%-24\%$ for promoted items \citep{gartner}. Stock-outs result in significant lost revenue for the supply chain. 
When a company faces a stock-out, roughly 70\% of customers do not wait for inventory to be replenished, but instead, purchase the items from a competitor \citep{bharadwaj2002retail}. 
Thus, in order to not lose customers and maximize profit, companies should have an inventory management system to provide high service level at a small cost.

Supply chains have different tools to balance between the service level and stock-out costs, and all of those tools use a kind of optimization or decision-making procedure.
For example, some companies produce huge products such as ships cannot hold inventory and have to balance their service level and service costs. 
Others can hold inventory and in this case the optimization problem finds a compromise between the holding and stock-out costs. 
Usually these models {\em assume} a given service level, and minimize the corresponding costs for that. 
As mentioned, the other relevant questions of an inventory management system are when, where, and how much of each item should be ordered, moved, stored, or transported.
These questions can be optimally answered by finding the optimal inventory policy for each node of the network. 

One category of models for multi-echelon inventory optimization is called the Stochastic Service Model (SSM) approach, which considers stochastic demand and stochastic lead times due to upstream stockouts. 
The optimal base-stock level can be found for serial systems without fixed costs by solving a sequence of single-variable convex problems \citep{clark1960optimal}. 
Similarly, by converting an assembly system (in which each node has at most one successor) to an equivalent serial system, the optimal solution can be achieved \citep{rosling1989optimal}. 
For more general network topologies, no efficient algorithm exists for finding optimal base-stock levels, and in some cases the form of the optimal inventory policy is not even known \citep{zipkin2000foundations}.

Another approach for dealing with multi-echelon problems is the Guaranteed Service Model (GSM) approach. 
GSM assumes the demand is bounded above, or equivalently the excess demand can be satisfied from outside of the system, e.g., by a third party vendor. It assumes a Committed Service Time (CST) for each node, which is the latest time that the node will satisfy the demand of its successor nodes. 
By this definition, instead of optimizing the inventory level, the GSM model optimizes the CST for each node, or equivalently it finds the base stock for each node of the network to minimize the holding costs.
This approach can handle more general supply chain topologies, typically using either dynamic programming \citep{graves1988safety,graves2000optimizing} or MIP techniques \citep{MagnantiShenShuSimchi-LeviTeo06}.
 
For a review of GSM and SSM Models see \cite{eruguz2016comprehensive},  \cite{simchi2011performance}, and \cite{snyder2018fundamentals}.

The sense among (at least some) supply chain practitioners is that the current set of inventory optimization models are sufficient to optimize most systems as they function normally. What keeps these practitioners up at night is the deviations from ``normal'' that occur on a daily basis and that pull the system away from its steady state. In other words, there is less need for new inventory optimization models and more need for tools that can help when the real system deviates from the practitioners' original assumptions. 

Our algorithm takes a snapshot of the supply chain at a given point in time and makes predictions about how individual components of the supply chain will perform, i.e., whether they will face stock-outs in the near future. We assume an SSM-type system, i.e., a system in which demands follow a known probability distribution, and stages within the supply chain may experience stock-outs, thus generating stochastic lead times to their downstream stages. The stages may follow any arbitrary inventory policy, e.g., base-stock or $(s,S)$. Classical inventory theory can provide long-term statistics about stock-out probabilities and levels (see, e.g., \cite{snyder2018fundamentals,zipkin2000foundations}), at least for certain network topologies and inventory policies. However, this theory does not make predictions about specific points in time at which a stock-out may occur. Since stock-outs are expensive, such predictions can be very valuable to companies so that they may take measures to prevent or mitigate impending stock-outs. 

Note that systems whose base-stock levels were optimized using the GSM approach may also face stock-outs, even though the GSM model itself assumes they do not.
The GSM approach assumes a bound on the demand value; when the real-world demand exceeds that bound, it may not be possible or desirable to satisfy the demand externally, as the GSM model assumes; therefore, stock-outs may occur in these systems. Therefore, stock-out prediction can be useful for fans of both SSM and GSM approaches. 

In a single-node network, one can obtain the stock-out probability and make stock-out predictions if the probability distribution of the demand is known (see Appendix \ref{sec:prdt_appd_one_agent_prediction}). 
However, to the best of our knowledge, there are no algorithms to provide stock-out predictions in multi-echelon networks. 
To address this need, in this paper, we propose an algorithm to provide stock-out predictions for each node of a multi-echelon network, which works for any network topology (as long as it contains no directed cycles) and any inventory policy.

The remainder of paper is organized as follows. In Section \ref{sec:prdt_solution_method}, we introduce our algorithm. Section \ref{sec:prdt_naive_approach} describes three naive algorithms to predict stock-outs.
To demonstrate the efficiency of the proposed algorithm in terms of solution quality, we compare our results with the best naive algorithms in Section \ref{sec:prdt_numerical_experiments}. Finally, Section \ref{sec:prdt_conclusions} concludes the paper and proposes future studies.

\section{Stock-out Prediction Algorithm}\label{sec:prdt_solution_method}

We develop an approach to provide stock-out predictions for multi-echelon networks with available data features. 
Our algorithm is based on deep learning, or a deep neural network (DNN). DNN is a non-parametric machine learning algorithm, meaning that it does not make strong assumptions about the functional relationship between the input and output variables. In the area of supply chain, DNN has been applied to demand prediction \citep{efendigil2009decision,DBLPjournalsCorrVieira15, ko2010review} and quantile regression \citep{taylor2000quantile,kourentzes2010advances,cannon2011quantile,xu2016quantile}.
It has also been successfully applied to the newsvendor problem with data features \citep{oroojlooyjadid2016applying}. 
The basics of deep learning are available in \cite{goodfellow2016deep}.

Consider a multi-echelon supply chain network with $n$ nodes, with arbitrary topology. For each node of the network, we know the history of the inventory level (IL), i.e., the on-hand inventory minus backorders, and of the inventory-in-transit (IT), i.e., the items that have been shipped to the node but have not yet arrived; the values of these quantities in period $i$ are denoted $IL_i$ and $IT_i$, respectively. In addition, we know the stock-out status for the node, given as a {\tt True} or {\tt False} Boolean, where {\tt True} indicates that the node experienced a stock-out. (We use 1 and 0 interchangeably with {\tt True} and  {\tt False}.) The historical stock-out information is not used to make predictions at time $t$ but is used to train the model.
The demand distribution can be known or unknown; in either case, we assume historical demand information is available.
The goal is to provide a stock-out prediction for each node of the network for the next period.

The available information that can be provided as input to the DNN algorithm includes the values of the $p$ available features (e.g., day of week, month of year, weather information), along with the historical observations of IL and IT at each node. 
Therefore, the available information for node $j$ at time $t$ can be written as:
%
\begin{equation}
[f_t^1, \dots, f_t^p, [IL^j_i, IT^j_i]_{i=1}^t],
\label{eq:prdt:observations}
\end{equation} 
where $f_t^1,\dots,f_t^p$ denotes the value of the $p$ features at time $t$. 

However, DNN algorithms are designed for inputs whose size is fixed; in contrast, the vector in \eqref{eq:prdt:observations} changes size at every time step. Therefore, we only consider historical information from the $k$ most recent periods instead of the full history. 
Although this omits some potentially useful information from the network, it unifies and reduces the input size, which has computational advantages, and selecting a large enough $k$ provides a good level of information about the system.  Therefore, the input of the DNN is:
\begin{equation}
[f_t^1, \dots, f_t^p, [IL_i, IT_i]_{i=t-k+1}^t]. \label{eq:inputs}
\end{equation}

The output of the DNN is the stock-out prediction for time $t+1$, for each node of the network, denoted $y_t = [y_t^1, \dots, y_t^n]$, a vector of length $n$.  
Each of the $y_t^j$, $j = 1, \cdots, n$, equals 1 if the node in period $t$ has stock-out and 0 otherwise.

A DNN is a network of nodes, beginning with an input layer (representing the inputs, i.e., \eqref{eq:inputs}), ending with an output layer (representing the $y_t$ vector), and one or more layers in between. Each node uses a mathematical function, called an activation function, to transform the inputs it receives into outputs that it sends to the next layer, with the ultimate goal of approximating the relationship between the overall inputs and outputs. \add{In a fully connected network, each node of each layer is connected to each node of the next layer through some coefficients}, called weights, which are initialized randomly. ``Training'' the network consists of determining good values for \add{those weights}, typically using nonlinear optimization methods. (A more thorough explanation of DNN is outside the scope of this paper; see, e.g., \cite{goodfellow2016deep}.) 

A loss function is used to evaluate the quality of a given set of weights. The loss function measures the distance between the predicted values and the known values of the outputs. We consider the following loss functions, which are commonly used for binary outputs such as ours:
\begin{itemize}
	\item Hinge loss function 
	\item Euclidean loss function 
	\item Soft-max loss function 
\end{itemize}
The hinge and Euclidean loss functions are reviewed in Appendix \ref{sec:prdt:appdx:losses}. The soft-max loss function uses the soft-max function, which is a generalization of logistic regression and is given by
\begin{equation} 
\label{eq:softmax_function}
\begin{split}
\sigma(z^u) = \frac{e^{z^u}}{\sum_{v=1}^U e^{z^v}} ;~~~~ \forall u = 1, \dots , U,
\end{split}
\end{equation} 
where $U$ is the number of possible categories (in our case, $U=2$), 
$$z^u =\sum \limits_{i=1}^{M_{L-1}} a_{i}^{L-1} w_{i,u},$$
$L$ is the number of layers in the DNN network, $a_i^{L-1}$ is the activation value of node $i$ in layer $L-1$, \add{$w_{i,u}$ is the weight between node $i$ in layer $L-1$ and node $u$ in layer $L$, }and $M_{L-1}$ represents the number of nodes in layer $L-1$.
Then the soft-max loss function is given by
\begin{equation} 
\label{eq:softmax_loss}
E =  
- \frac{1}{M} \sum \limits_{i=1}^M \sum \limits_{u=1}^U {\mathbb I} \{y_i = u-1 \} \log \frac{e^{z^u_i}}{\sum_{v=1}^U e^{z^v_i}},
\end{equation} 
where $M$ is the total number of training samples, ${\mathbb I}(\cdot)$ is the indicator function, and $E$ is the loss function value, which evaluates the quality of a given classification (i.e., prediction). In essence, the loss function \eqref{eq:softmax_loss} penalizes predictions $y_i$ that differ from the value given by the loss function \eqref{eq:softmax_function}.

The hinge and soft-max function 
provide a probability distribution over $U$ possible classes; we then take the {\tt argmax} over them to choose the predicted class. 
In our case there are $U=2$ classes, i.e., {\tt True} and {\tt False} values, as required in the prediction procedure.
On the other hand, the Euclidean function 
provides a continuous value, which must be changed to a binary output. In our case, we round Euclidean loss function values to their nearest value, either 0 or 1. 

Choosing weights for the neural network involves solving a nonlinear optimization problem whose objective function is the loss function and whose decision variables are the network weights. Therefore, we need gradients of the loss function with respect to the weights; these are usually obtained using back-propagation or automatic differentiation. The weights are then updated using a first- or second-order algorithm, such as gradient descent, stochastic gradient descent (SGD), SGD with momentum, LBFGS, etc. Our procedure repeats iteratively until one of the following stopping criteria is met:
\begin{itemize}
\item The loss function value is less than {\tt Tol} 
\item The number of passes over the training data reaches {\tt MaxEpoch}
\end{itemize}
{\tt Tol} and {\tt MaxEpoch} are parameters of the algorithm; we use {\tt Tol}$=1e-6$ and {\tt MaxEpoch$=3$}

The loss function provides a measure for monitoring the improvement of the DNN algorithm through the iterations. 
However, it cannot be used to measure the quality of prediction, and it is not meaningful by itself.
Since the prediction output is a binary value, the test error---the number of wrong predictions divided by the number of samples---is an appropriate measure.
Moreover, statistics on false positives (type I error, the incorrect rejection of a true null hypothesis) and false negatives (type II error, the failure to reject a true null hypothesis) are helpful, and we use them to get more insights about how the algorithm works.

The DNN algorithm provides one prediction, in which the false positive and negative errors are weighted equally.
However, the modeler should be able to control the likelihood of a stock-out prediction, i.e., the balance between  false positive and false negative errors. 
To this end, we would benefit from a loss function that can provide control over the likelihood of a stock-out prediction, since the DNN's output is directly affected by its loss function. 

The loss functions mentioned above do not have any weighting coefficient, and place equal weight between selecting 0 (predicting no stock-out) and 1 (predicting stock-out). 
To correct this, we propose weighing the loss function value that is incurred for each output, 0 and 1, using weights $c_n$ and $c_p$, which represent the costs of false positive and negative errors, respectively. 
In this way, when $c_p < c_n$, the DNN tries to have a smaller number of cases in which it returns {\tt False} 
but in fact $y_i = 0$, so it predicts more stock-outs to result in a smaller number of false negative errors and a larger number of false positive errors. Similarly, when $c_p > c_n$, the DNN predicts fewer stock-outs to avoid cases in which it returns {\tt True} but in fact $y_i = 1$. Therefore, it makes a smaller number of false positive errors and a larger number of false negative errors. 
If $c_n = c_p$, our revised loss function works similarly to the original loss functions.

Using this approach, the weighted hinge, Euclidean, and soft-max  loss functions are as follows.
%

Hinge:
\begin{subequations}\label{prdt:eq:hinge_loss_weighted}
	\begin{align}
	E &= \frac{1}{N} \sum \limits_{i=1}^N E_i \label{prdt:eq:hinge_loss_weighted_0} \\
	E_i &= \begin{cases}
	c_n \max(0, 1 - y_i \hat{y}_i) ~, \text{if}~~ y_i = 0 \\
	c_p \max(0, 1 - y_i \hat{y}_i) ~, \text{if}~~ y_i = 1,
	\end{cases} \label{prdt:eq:hinge_loss_weighted_1}
	\end{align}
\end{subequations}

Euclidean:
\begin{subequations}\label{prdt:eq:euclidean_loss_weighted}
	\begin{align}
	E &= \frac{1}{N} \sum \limits_{i=1}^N E_i \label{prdt:eq:euclidean_loss_weighted_0} \\
	E_i &= \begin{cases}
	c_n ||y_i - \hat{y}_i ||_2^2 ~, \text{if}~~ y_i = 0\\
	c_p ||y_i - \hat{y}_i ||_2^2 ~, \text{if}~~ y_i = 1, 
	\end{cases} \label{prdt:eq:euclidean_loss_weighted_1}
	\end{align}
\end{subequations}

Soft-max:
\begin{equation} 
\label{prdt:eq:softmax_loss_weighted}
E =  
- \frac{1}{N} \sum \limits_{i=1}^N \sum \limits_{u=1}^U w_u {\mathbb I} \{y_i = u-1 \} \log \frac{e^{z^u_i}}{\sum_{v=1}^U e^{z^v_i}},
\end{equation} 
where $U=2$, $w_1 = c_n$, and $w_2 = c_p$. 
Thus, these loss functions allow one to manage the number of false positive and negative errors.

\section{Naive Approaches}\label{sec:prdt_naive_approach}

In this section, we propose three naive approaches to predict stock-outs. These algorithms are used as baselines for measuring the quality of the DNN algorithm. They are easy to implement, but they do not consider the system state at any nodes other than the node for which we are predicting stockouts. (The proposed DNN approach, in contrast, uses the state at all nodes to provide a more effective prediction.) 

In all of the naive algorithms, we use $IP_t$ to denote the inventory position in period $t$. 
Also, $v$ and $u$ are the numbers of the training and testing records, respectively, and $d = [ d_1 , d_2 , \cdots,d_v]$ is the demand of the customers in each period of the training set.
Finally, 
the function {\tt approximator($s$)} takes a list $s$ of numbers, fits a normal distribution to it, and returns the corresponding parameters of the normal distribution.

{   
\singlespacing
%
\begin{algorithm}[]
	\caption{Naive Algorithm 1}
	\label{Naive_1}
	\begin{algorithmic}[1]
		\Procedure{Naive-1} {}
		\State given $\alpha$ as an input;
		\State $s = \{IP_t|y_{t+1}=1, t=1,\dots,v\}$; \Comment {Training procedure}		
		\State $\mu_{s} , \sigma_s$ = {\tt approximator($s$)};
		\State $\eta_\alpha = \mu_{s} + \Phi^{-1}_\alpha (\sigma_s)$;		
		\For{$t=1 : u$}		\Comment{Testing procedure}
		\If {$IP_t < \eta_\alpha$} 
		\State  prediction$(t) = 1;$		
		\Else
		\State  prediction$(t) = 0;$	
		\EndIf
		\EndFor
		\State \Return prediction
		\EndProcedure
	\end{algorithmic}
\end{algorithm}
} 

Naive Algorithm \ref{Naive_1} first determines all periods in the training data in which a stock-out occurred and builds a list $s$ of the inventory positions in the preceding period for each. Then it fits a normal distribution $ {\cal{N}} (\mu_{s}, \sigma_{s})$ to the values in $s$ and calculates the $\alpha$th quantile of that distribution, for a given value of $\alpha$. Finally, it predicts a stock-out in period $t+1$ if $IP_t$ is less than that quantile. The value of $\alpha \in (0,1)$ is determined by the modeler. 

Naive Algorithm \ref{Naive_2} groups the inventory positions into a set of ranges, calculates the frequency of stock-outs in the training data for each range, and then predicts a stock-out in period $t+1$ if the range that $IP_t$ falls into experienced stock-outs is $\gamma$ times more than of the time in the training data.

Finally, Naive Algorithm \ref{Naive_3} uses classical inventory theory, which says the inventory level in period $t+L$ equals $IP_t$ minus the lead-time demand, where $L$ is the lead time \citep{zipkin2000foundations,snyder2018fundamentals}. 
The algorithm estimates the lead-time demand distribution by fitting a normal distribution based on the training data, then predicts a stockout in period $t+1$ if $IP_t$ is less than or equal to the $\alpha$th quantile of the estimated lead-time demand distribution, where $\alpha$ is a parameter chosen by the modeler.

{   
\singlespacing
\begin{algorithm}[]
	\caption{Naive Algorithm 2}
	\label{Naive_2}
	\begin{algorithmic}[1]
		\Procedure{Naive-2} {}
		\State $l = \min_{t=1}^v\{IP_t\}$; $u = \max_{t=1}^v\{IP_t\}$; 
		\State given $\gamma$ as an input;
		\State Divide $[l,u]$ into $k$ equal intervals $[l_i,u_i], \forall i=1,\cdots,k$;
		\State $SO_i = NSO_i = 0$ $\forall i=1,\cdots,k$;
		\For{$t=1 : v$} \Comment{Training procedure}
		\State {$s(t) = i$ such that $IP_t \in [l_i,u_i]$}
		\If {$y_{t+1} =$ 1} 
		\State $SO_{s(t)}$ $+= 1;$
		\Else
		\State $NSO_{s(t)}$ $+= 1;$		
		\EndIf
		\EndFor
		\For{$t=1 : u$}		\Comment{Testing procedure}
		\State {$s(t) = i$ such that $IP_t \in [l_i,u_i]$}
		\If {$SO_{s(t)} *\gamma > NSO_{s(t)} $} 
		\State prediction$(t) = 1;$		
		\Else
		\State prediction$(t) = 0;$	
		\EndIf
		\EndFor
		\State \Return prediction
		\EndProcedure
	\end{algorithmic}
\end{algorithm}

\begin{algorithm}[]
	\caption{Naive Algorithm 3}
	\label{Naive_3}
	\begin{algorithmic}[1]
		\Procedure{Naive-3} {}
		\State $\mu_{d} , \sigma_d$ = {\tt approximator($\{d_t\}_{t=1}^v$)}; \Comment{Training procedure}
		\State given $\alpha$ as an input;
		\State $\eta_\alpha = \mu_{d} + \Phi^{-1}_\alpha (\sigma_d)$;
		\For{$t=1 : u$}		\Comment{Testing procedure}
		\If {$IP_t < \eta_\alpha$} 
		\State prediction$(t) = 1;$		
		\Else
		\State prediction$(t) = 0;$	
		\EndIf
		\EndFor
		\State \Return prediction
		\EndProcedure
	\end{algorithmic}
\end{algorithm}
}

In Naive Algorithms \ref{Naive_1} and \ref{Naive_3}, the value of $\alpha$ (and hence $\eta_\alpha$) \add{and the value of $\gamma$ in Naive Algorithm \ref{Naive_2}} is selected by the modeler. 
A small value of $\alpha$ results in a small $\eta_\alpha$ so that the algorithm predicts fewer stock-outs. \add{The same is true for a small $\gamma$.}
Generally, as $\alpha$ \add{ or $\gamma$} decreases, the number of false positive errors decreases compared to the number of false negative errors, and vice versa.
Thus, selecting an appropriate value of $\alpha$ \add{or $\gamma$} is important and directly affects the output of the algorithm. 
Indeed, the value of $\alpha$ \add{or $\gamma$} has to be selected according to the preferences of the company running the algorithm. 
For example, a company may have very expensive stock-outs. 
So, it may choose a very large $\alpha$ \add{or $\gamma$} so that the algorithm predicts frequent  stock-outs, along with many more false positive errors, and then checks them one by one to prevent the stock-outs. 
In this situation the number of false positive errors increases; however, the company faces fewer false negative errors, which are costly. 
In order to determine an appropriate value of $\alpha$ \add{or $\gamma$}, the modeler should consider \add{the costs of false positive and negative errors, i.e., $c_p$ and $c_n$, respectively}. 


\section{Numerical Experiments}\label{sec:prdt_numerical_experiments}

\begin{figure}[]
	\centering		
	\includegraphics[scale=1.3]{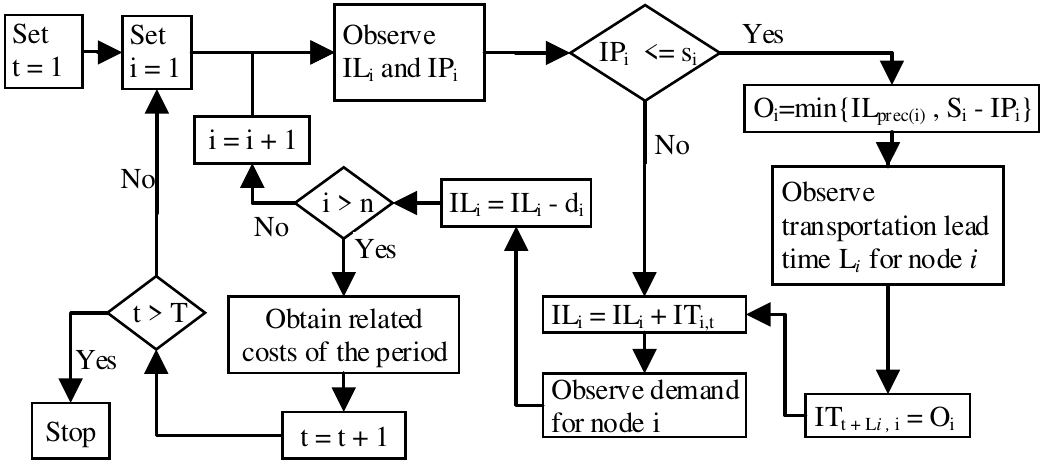}
	\caption{The simulation algorithm used to simulate a supply network}
	\label{prdt_simulationFlowchart}
\end{figure}

In order to check the validity and accuracy of our algorithm, we conducted a series of numerical experiments. 
Since there is no publicly available data of the type needed for our algorithm, we built a simulation model that assumes each node follows a base-stock policy and can make an order only if its predecessor has enough stock to satisfy it so that only the retailer nodes face stock-outs.
The simulation records several state variables for each of the $n$ nodes and for each of the $T$ time periods.
Figure \ref{prdt_simulationFlowchart} shows the flowchart of the simulation algorithm used. 

To see how our algorithm works with different network topologies, we conducted multiple tests on five supply chain network topologies, ranging from a simple series system to complex networks containing (undirected) cycles and little or no symmetry. 
These tests are intended to explore the robustness of the DNN approach on simple or very complex networks. The five supply chain networks we used are: 
\begin{itemize}
	\item Serial network with 11 nodes.
	\item One warehouse, multiple retailer (OWMR) network with 11 nodes.
	\item Distribution network with 13 nodes.
	\item Complex network~I with 11 nodes, including one retailer and two warehouses.
	\item Complex network~II with 11 nodes, including three retailers and one node at the farthest echelon upstream (which we refer to as a warehouse).	
\end{itemize}
%

\begin{figure}
	\centering{\includegraphics[scale=0.4]{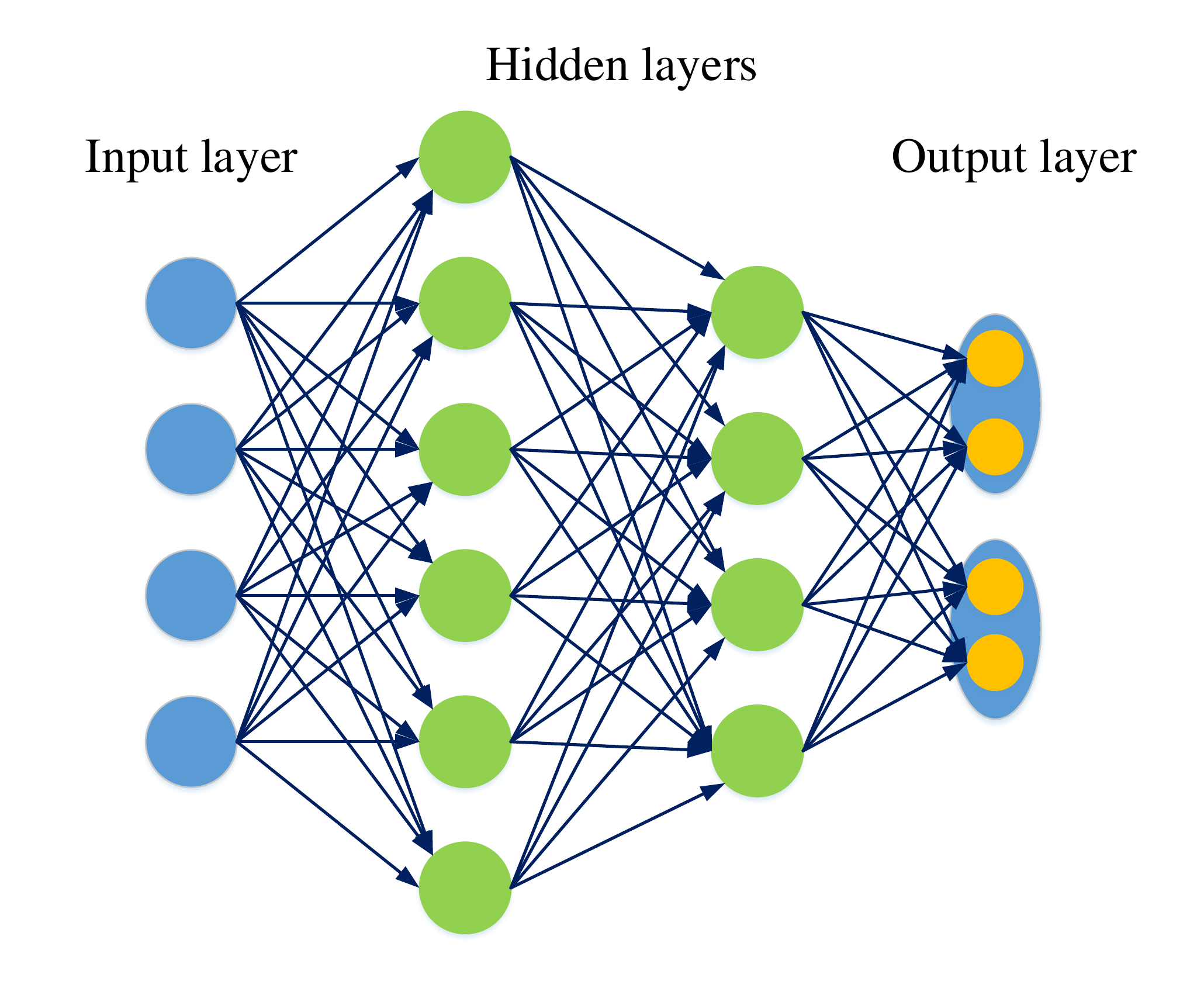}}
	\caption{A network used to predict stock-outs of two nodes. For each of the networks, we used a similar network with $n$ soft-max outputs.}
	\label{fig:dnn_softmax}
\end{figure}

We simulated each of the networks for $10^6$ periods,  with 75\% of the resulting data used for training (and validation) and the remaining 25\% for testing.
For all of the problems we used a fully connected DNN network with 350 and 150 sigmoid nodes in the first and second layers, respectively. 
The inputs are the inventory levels and on-order inventories for each node from each of the \add{$k=11$} most recent periods \add{(as given in \eqref{eq:inputs})}, and the output is the binary stock-out predictor for each of the nodes. Figure \ref{fig:dnn_softmax} shows a general view of the DNN network. 
Among the loss functions reviewed in Section \ref{sec:prdt_solution_method}, the soft-max loss function had the best accuracy in initial numerical experiments. 
Thus, the soft-max loss function was selected and its results are provided.
To this end, we implemented the weighted soft-max function and its gradient (see Appendix \ref{sec:prdt_appd_grdt_weighted_softmax})  in the DNN computation framework Caffe \citep{jia2014caffe}, and all of the tests were done on machines with 16 AMD cores and 32 GB of memory. 
In order to optimize the network, the SGD algorithm---with batches of 50---with momentum is used, and each problem is run with \add{{\tt MaxEpoch=3}}. Each epoch defines one pass over the training data. 
Finally, we tested 99 values of $\alpha \in \{0.01, 0.02, \dots, 0.99\}$ and 118 values of $(c_n,c_p)$, such that $\{c_p, c_n\} \in [0.3,15]$.

The DNN algorithm is scale dependent, meaning that the algorithm hyper-parameters (such as $\gamma$, learning rate, momentum, etc.; see \cite{goodfellow2016deep})~are dependent on the values of $c_p$ and $c_n$.  
Thus, a set of appropriate hyper-parameters of the DNN network for a given set of cost coefficients $(c_p,c_n)$ does not necessarily work well for another set $(c'_p,c'_n)$. 
This means that, ideally, for each set of $(c_p,c_n)$, we should re-tune the DNN hyper-parameters, i.e., re-train the network.
However, the tuning procedure is computationally expensive, so in our experiments we tuned the hyper-parameters for $c_p = 2$ and $c_n = 1$ and used the resulting value for other sets of costs, in all network topologies. 
However, in complex network II, we did not get good convergence using this method, so we tuned the network for another set of cost coefficients to make sure that we get a not-diverging DNN for each set of coefficients. To summarize, our experiments use minimal tuning, except for complex network II---still far less than the total numbers of possible experiment---, but even so, the algorithm performs very well; however, better tuning could further improve our results.

In what follows, we demonstrate the results of the DNN and three naive algorithms in seven experiments.
Sections \ref{sec:prdt_section_result_serial}--\ref{sec:prdt_section_result_complexII} present the results of the serial, OWMR, distribution, complex I, and complex II networks, respectively. Section \ref{sec:prdt:extentions} extends these experiments: Section \ref{sec:prdt:extebded_result_ind10} provides threshold prediction,  Section \ref{sec:prdt:result_dependent_demand} analyzes the results of a distribution network with multiple items with dependent demand, and Section \ref{sec:prdt:result_multi_period} shows the results of predicting stock-outs multiple periods ahead  in a distribution network.
In each of the network topologies, we plot the false positive vs.~false negative errors for all algorithms to compare their performance.
In addition, two other figures in each section show the accuracy vs.~false positive and negative errors to provide better insights into the way that the DNN algorithm (weighted and unweighted) works compared to the naive algorithms.

\subsection{Results: Serial Network}\label{sec:prdt_section_result_serial}

\begin{figure}[]
	\centering
	\caption{The serial network}
	\label{fig:serial-1-10}
	\includegraphics[scale=0.7]{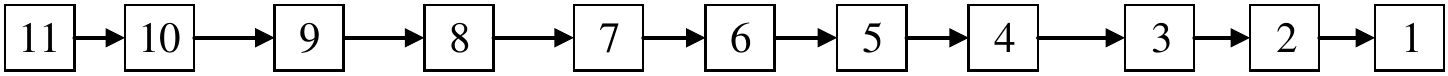}
\end{figure}

\begin{figure}[H]
	\centering
	\includegraphics[scale=0.55]{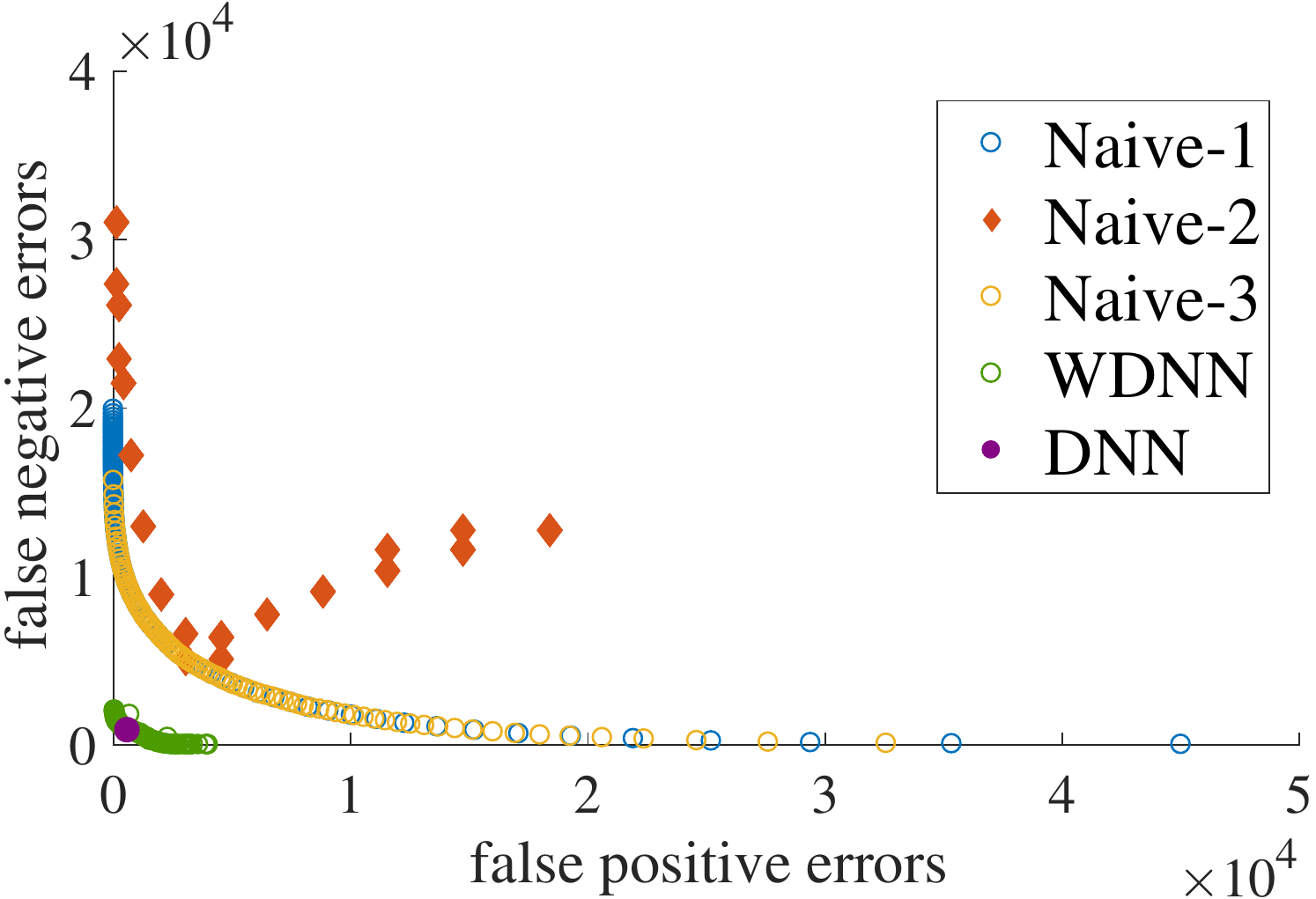}
	\caption{False positives vs.~false negatives for the serial network}
	\label{fig:prdt_serial-weighted-results}
\end{figure}

\begin{figure}[H]
	\centering
	\includegraphics[scale=0.4]{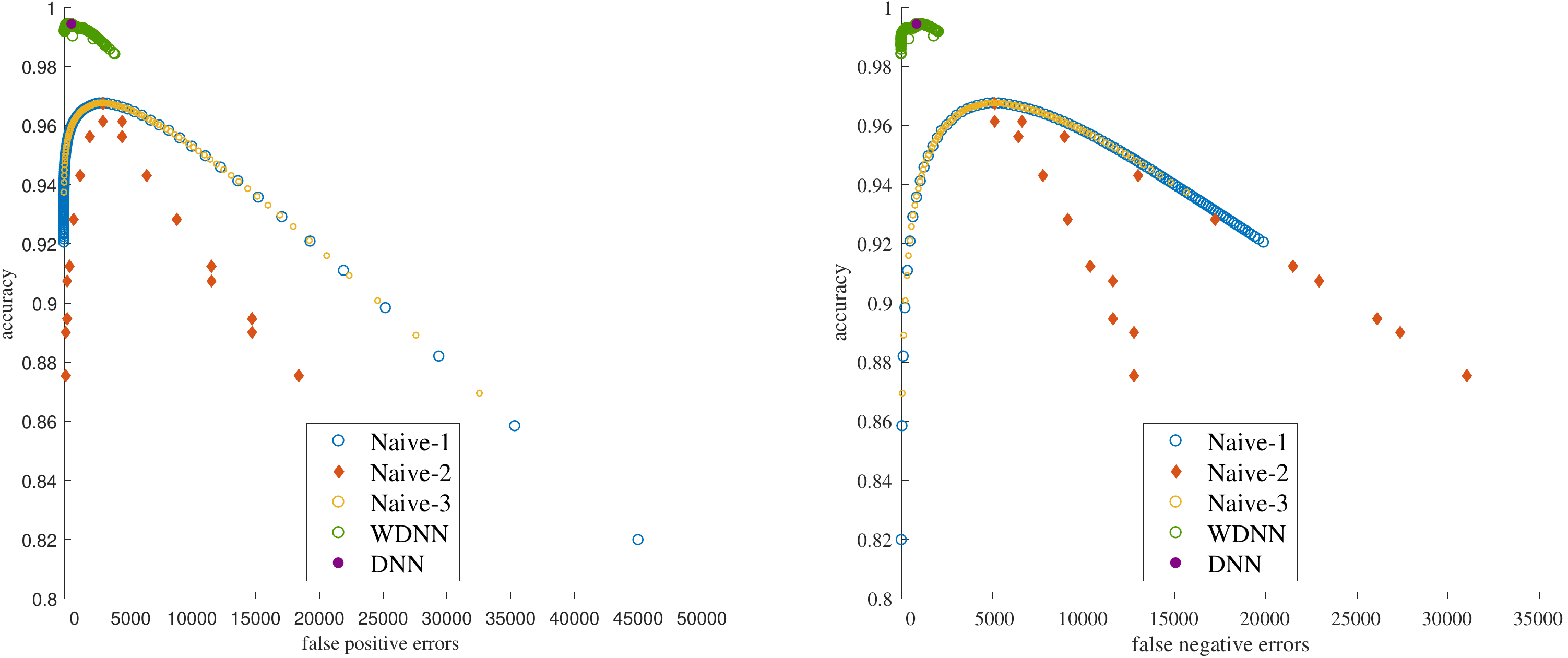}
	\caption{Accuracy of each algorithm for the serial network}
	\label{fig:prdt_serial-accuracy}
\end{figure}

Figure \ref{fig:serial-1-10} shows the serial network with 11 nodes.
The training dataset is used to train all five algorithms and the corresponding results are shown in Figures~\ref{fig:prdt_serial-weighted-results} and \ref{fig:prdt_serial-accuracy}.  
Figure~\ref{fig:prdt_serial-weighted-results} plots the false-negative errors vs.~the false-positive errors for each approach and for a range of $\alpha$ values for the naive approaches and a range of weights for the weighted DNN approach. 
Points closer to the origin indicate more desirable solutions.
Since there is just one retailer, 
the algorithms each make $2.5\times 10^5$ stock-out predictions (one in each of the $2.5\times 10^5$ testing periods); therefore, the number of errors in both figures should be compared to $2.5\times 10^5$. 

The DNN approach always dominates the naive approaches, with the unweighted version providing a slightly better accuracy but the weighted version providing more flexibility. 
For any given number of false-positive errors, the numbers of false-negative errors of the DNN and WDNN algorithms are smaller than those of the naive approaches, and similarly for a given number of false-negative errors. 
The results of naive approaches are similar to each other, with Naive-1 and Naive-3 outperforming Naive-2 for most $\alpha$ values.
Similarly, Figure~\ref{fig:prdt_serial-accuracy} plots the errors vs.~the accuracy of the predictions and shows that for a given number of false positives or negatives, the DNN approaches attain a much higher level of accuracy than the naive approaches do. 
In conclusion, the naive algorithms perform similar to each other and worse than DNN, since they do not use the available historical information. In contrast, DNN learns the relationship between state inputs and stock-outs and can predict stock-outs very well.

\subsection{Results: OWMR Network}\label{sec:prdt_section_result_tree}

\begin{figure}[H]
	\centering
	\includegraphics[scale=0.7]{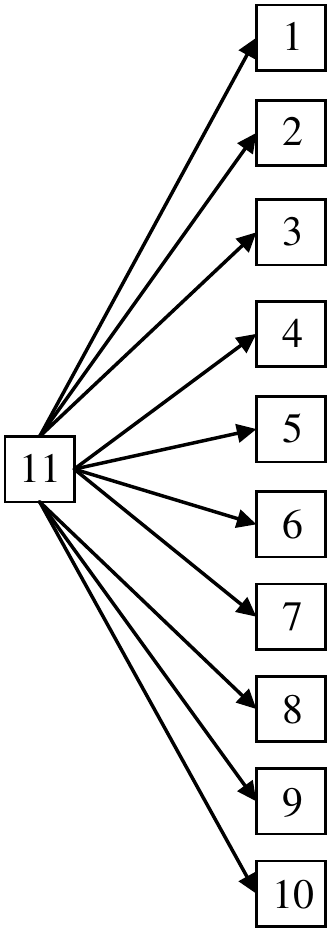}
	\caption{The OWMR network}
	\label{fig:prdt_tree-1-10}
\end{figure}

\begin{figure}[H]
	\centering
	\includegraphics[scale=0.55]{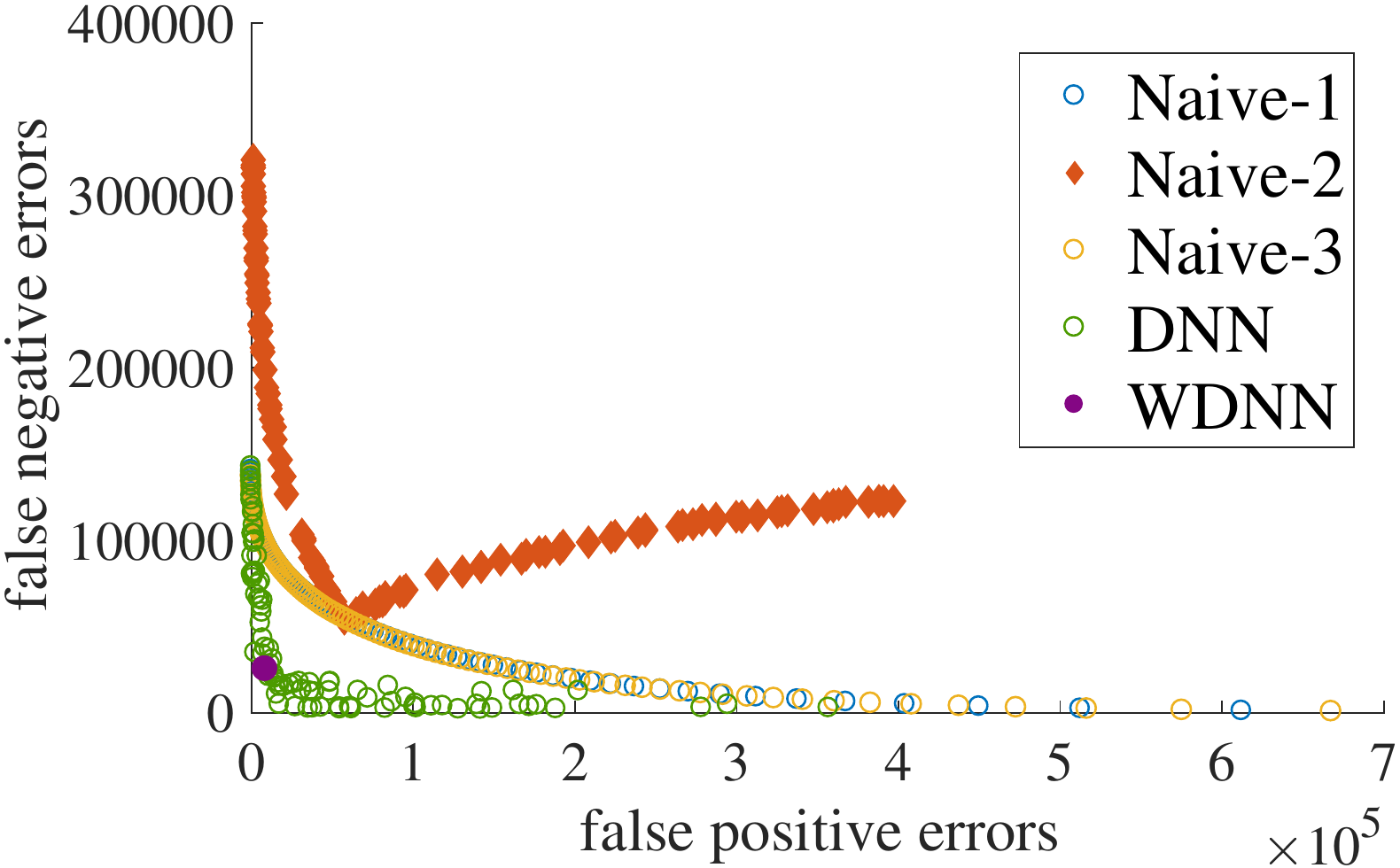}
	\caption{False positives vs.~false negatives for the OWMR network}
	\label{fig:prdt_tree-results}
\end{figure}

\begin{figure}[H]
	\centering
	\includegraphics[scale=0.4]{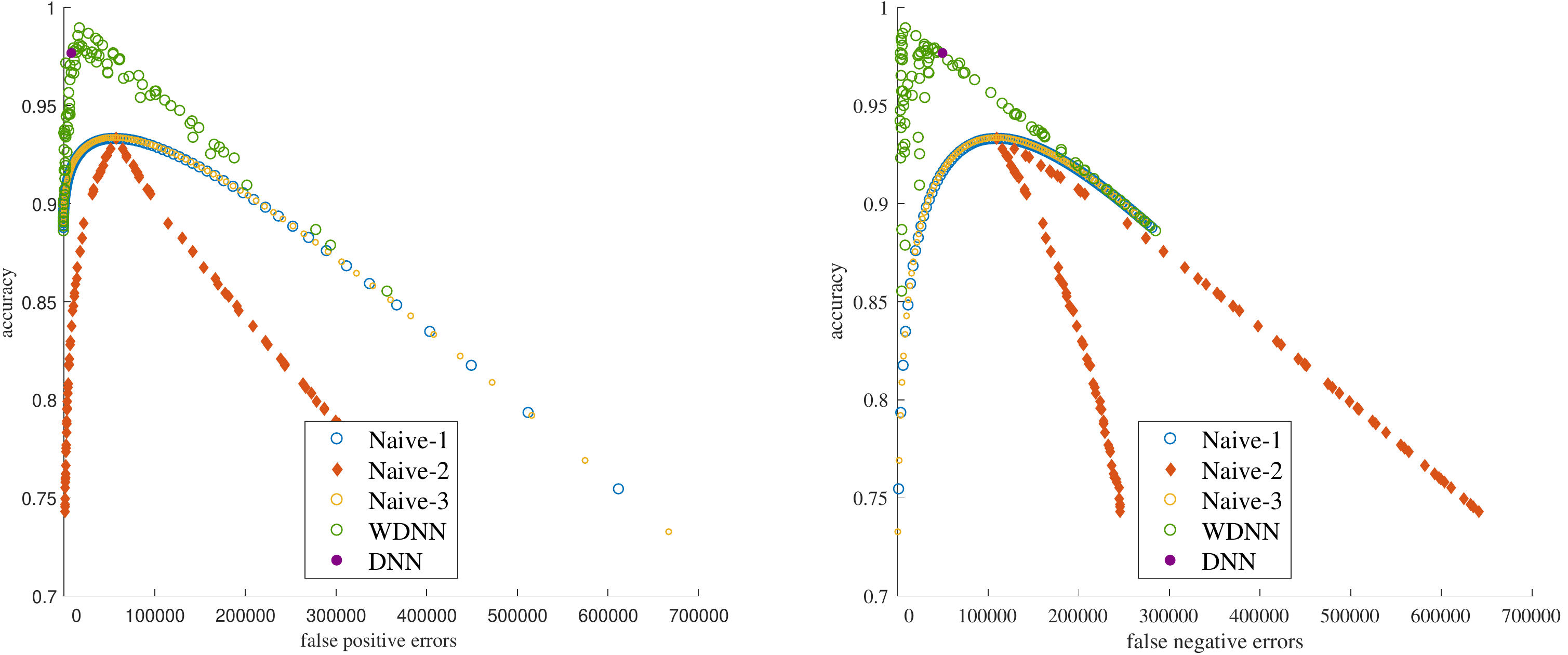}
	\caption{Accuracy of each algorithm for the OWMR network}
	\label{fig:prdt_tree-accuracy}
\end{figure}

Figure \ref{fig:prdt_tree-1-10} shows the OWMR network with 11 nodes and Figures \ref{fig:prdt_tree-results} and  \ref{fig:prdt_tree-accuracy} present the experimental results for this network.
Since there are 10 retailers, prediction is more challenging than for the serial network, as the algorithms each make $2.5\times 10^6$ stock-out predictions; the number of errors in both figures should be compared to $2.5\times 10^6$. 

Figure~\ref{fig:prdt_tree-results} shows the false-negative errors vs.~the false-positive errors for each approach and for a range of $\alpha$ values for the naive approaches and a range of weights for the weighted DNN approach. 
DNN and weighted DNN dominate the naive approaches.
The three naive approaches are similar to each other, with Naive-2 somewhat worse than the other two.
Figure~\ref{fig:prdt_tree-accuracy} plots the errors vs. the accuracy of the predictions and confirms that DNN can attain higher accuracy levels for the same number of errors than the naive approaches.
It is also apparent that all methods are less accurate for the OWMR system than they are for the serial system since there are many more predictions to make. However, DNN still provides better accuracy compared to the naive approaches.

\subsection{Results: Distribution Network}\label{sec:prdt_section_result_distribution}
\begin{figure}[H]
	\centering
	\includegraphics[scale=1]{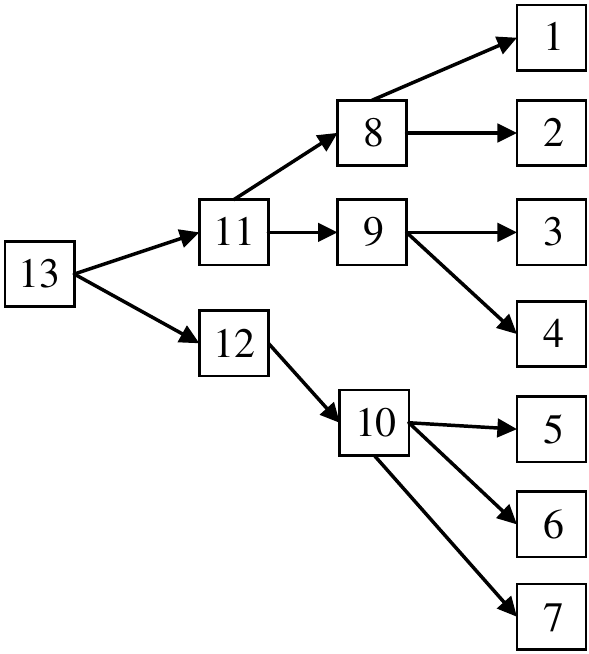}
	\caption{The distribution network}
	\label{fig:prdt_distribution-1-2-3-7}
\end{figure}

\begin{figure}[H]
	\centering
	\includegraphics[scale=0.55]{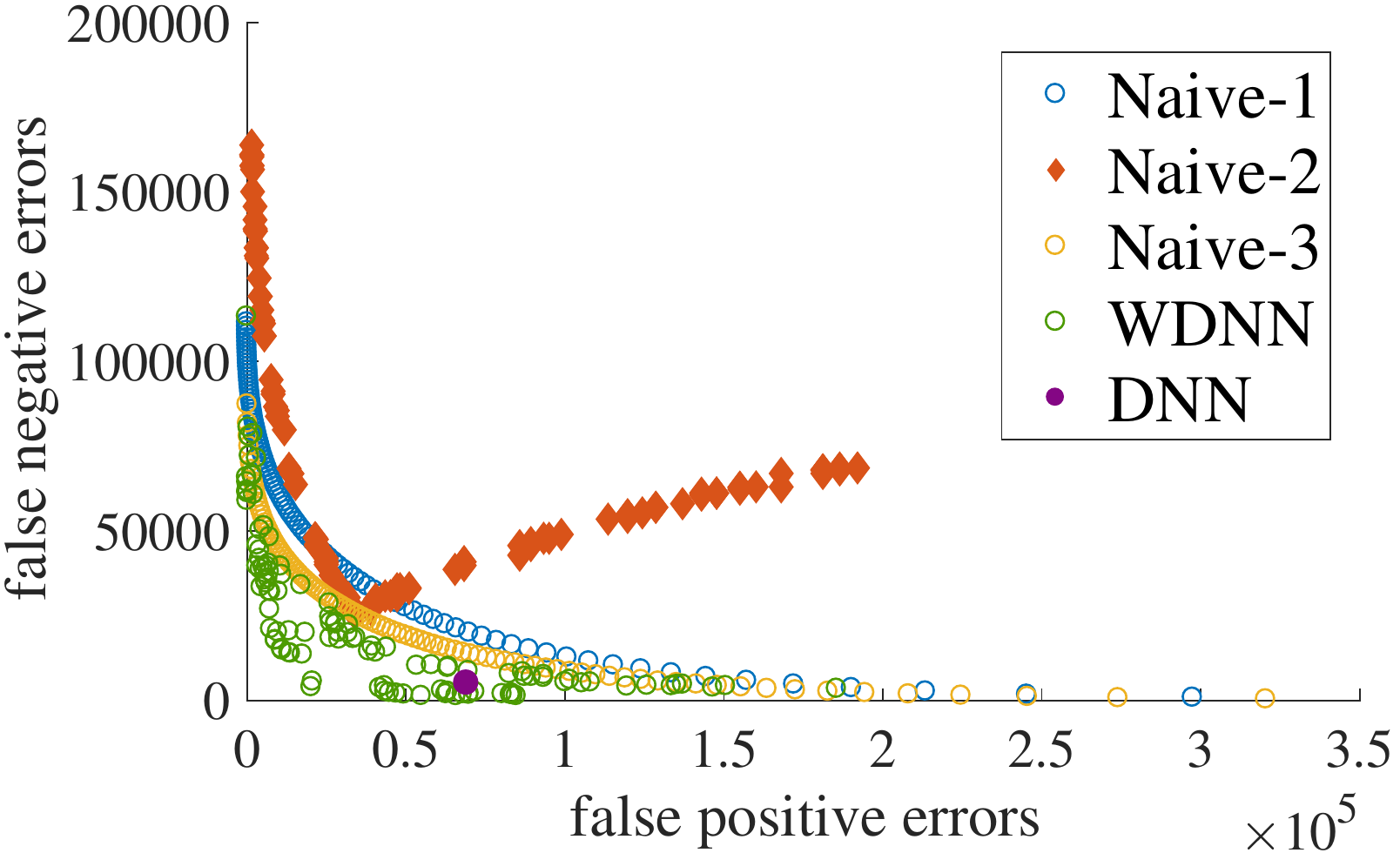}
	\caption{False positives vs.~false negatives for the distribution network}
	\label{fig:prdt_distribution_weighted_DNN_Naive}
\end{figure}

Figure \ref{fig:prdt_distribution-1-2-3-7} shows the distribution network with 13 nodes, and Figure~\ref{fig:prdt_distribution_weighted_DNN_Naive} 
provides the corresponding results of the five algorithms. 
As Figure~\ref{fig:prdt_distribution_weighted_DNN_Naive} shows, the DNN approach mostly dominates the naive approaches. However,it does not perform as well as in serial or OWMR networks; that occurs because of the tuning of the DNN network hyper-parameters.
Among the three naive approaches, Naive-3 dominates Naive-1, since the demand data comes from a normal distribution without any noise, and the algorithm also approximates a normal distribution, which needs around 12 samples to get a good estimate of the mean and standard deviation.
Therefore, the experiment is biased in favor of Naive-3. 
Plots of the errors vs. the accuracy of the predictions are similar to those in Figure~\ref{fig:prdt_tree-accuracy}; they are omitted to save space.

Compared to the OWMR network, the distribution network includes fewer retailer nodes and therefore fewer stock-out predictions; however, the network is also more complex, and as a result the DNN is less accurate than it is for OWMR network. We conclude that the accuracy of the DNN depends more on the number of echelons in the system than it does on the number of retailers. 
On the other hand, DNN obtains greater accuracy than any of the naive approaches.

\subsection{Results: Complex Network I}\label{sec:prdt_section_result_complexI}

\begin{figure}[H]
	\centering
	\includegraphics[scale=1]{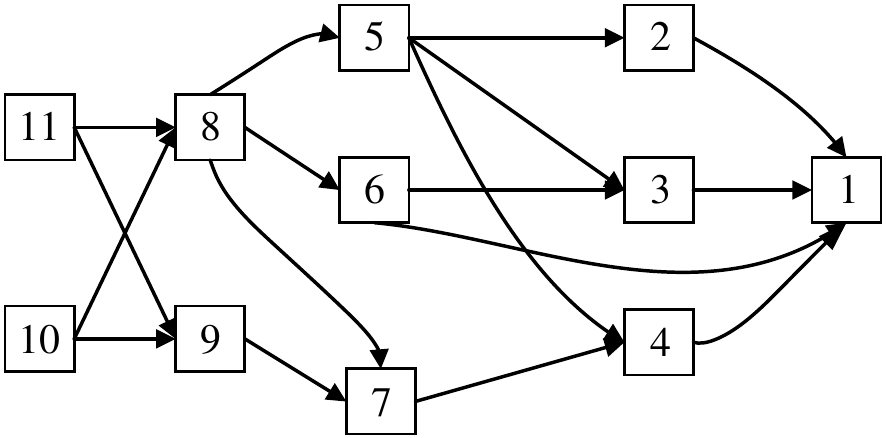}
	\caption{The complex network, two warehouses}
	\label{fig:prdt_complex-2-2-3-3-1}
\end{figure}

\begin{figure}[H]
	\centering
	\includegraphics[scale=0.55]{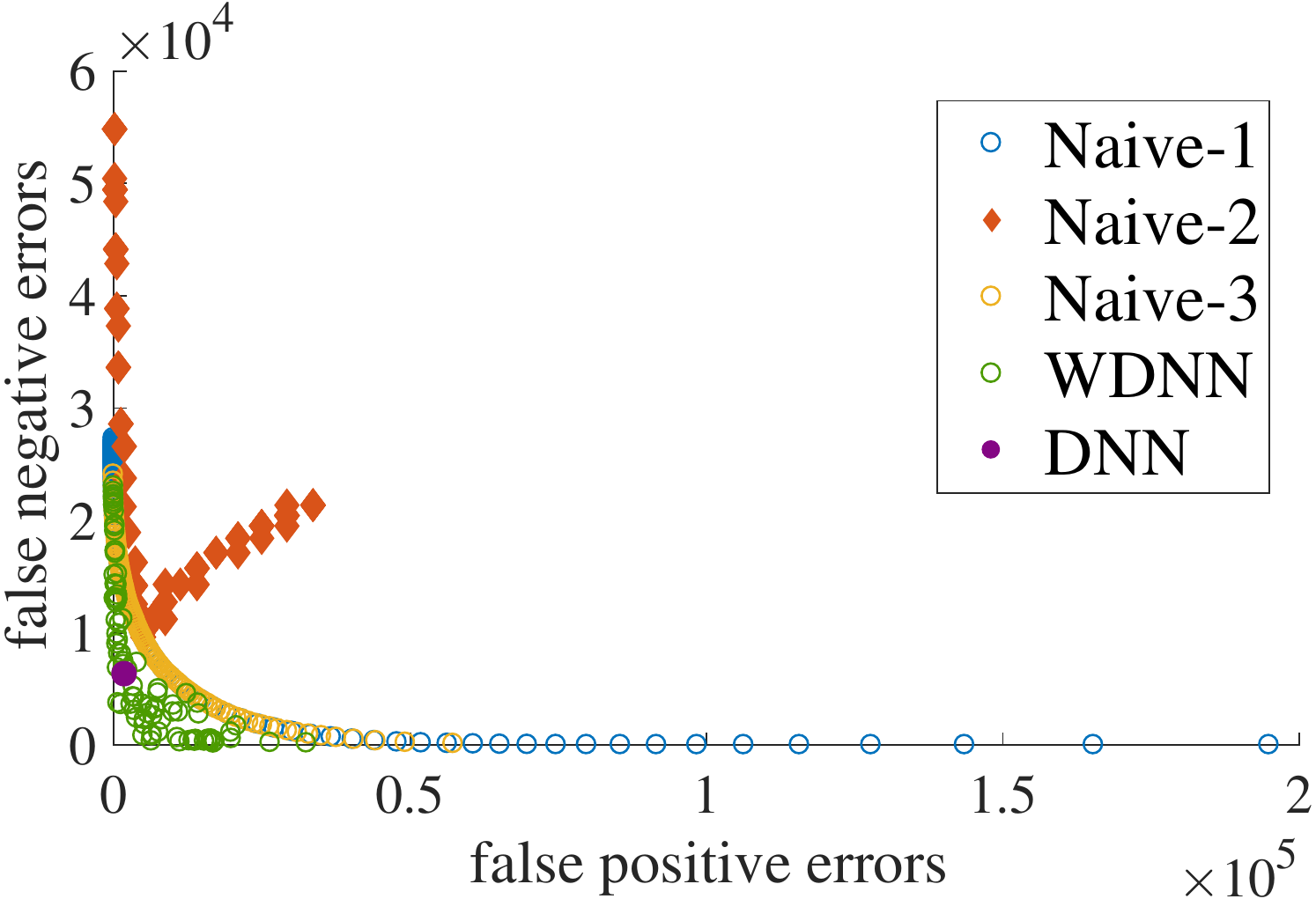}
	\caption{False positives vs.~false negatives for complex network I}
	\label{fig:prdt_complex_2_2-accuracy}
\end{figure}

Figure \ref{fig:prdt_complex-2-2-3-3-1} shows a complex network with two warehouses (i.e., two nodes at the farthest echelon upstream), and Figure~\ref{fig:prdt_complex_2_2-accuracy} 
presents the corresponding results of the five algorithms.  
Figure~\ref{fig:prdt_complex_2_2-accuracy} plots the false-negative errors vs.~the false-positive errors for each approach and for a range of $\alpha$ values for the naive approaches and a range of weights for the weighted DNN approach. 
The DNN approach dominates the naive approaches for most cases, but does worse when false-positives are tolerated in favor of reducing false-negatives. 
The average accuracy rates for this system are 91\% for WDNN and 97\% for DNN, which show the importance of hyper-parameter tuning for each  weight of the weighted DNN approach. 
Tuning it for each weight individually would improve the results significantly (but increase the computation time). 
Plots of the errors vs. the accuracy of the predictions are similar to those in Figure~\ref{fig:prdt_tree-accuracy}; they are omitted to save space.
	
As in the serial network, there is just one retailer node; however, since the network is more complex, DNN produces less accurate predictions for complex network I than it does for the serial network, or for the other tree networks (OWMR and distribution). The added complexity of this network topology has an effect on the accuracy of our model, though the algorithm is still quite accurate.

\subsection{Results: Complex Network II}\label{sec:prdt_section_result_complexII}

\begin{figure}[H]
	\centering
	\includegraphics[scale=1]{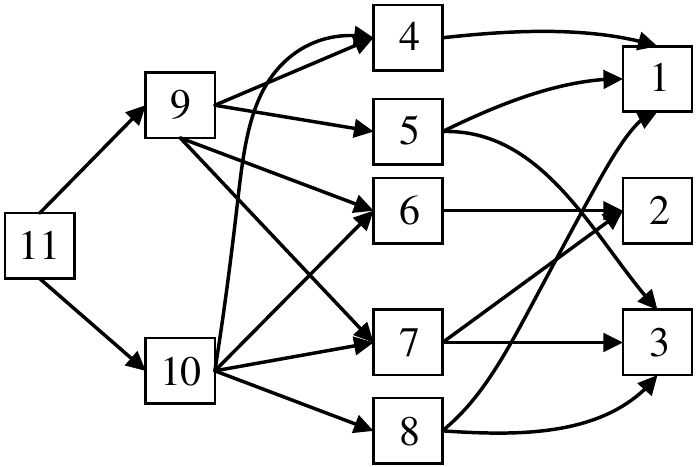}
	\caption{The complex network, three retailers}
	\label{fig:prdt_complex-1-2-5-3}
\end{figure}

\begin{figure}[H]
	\centering
	\includegraphics[scale=0.55]{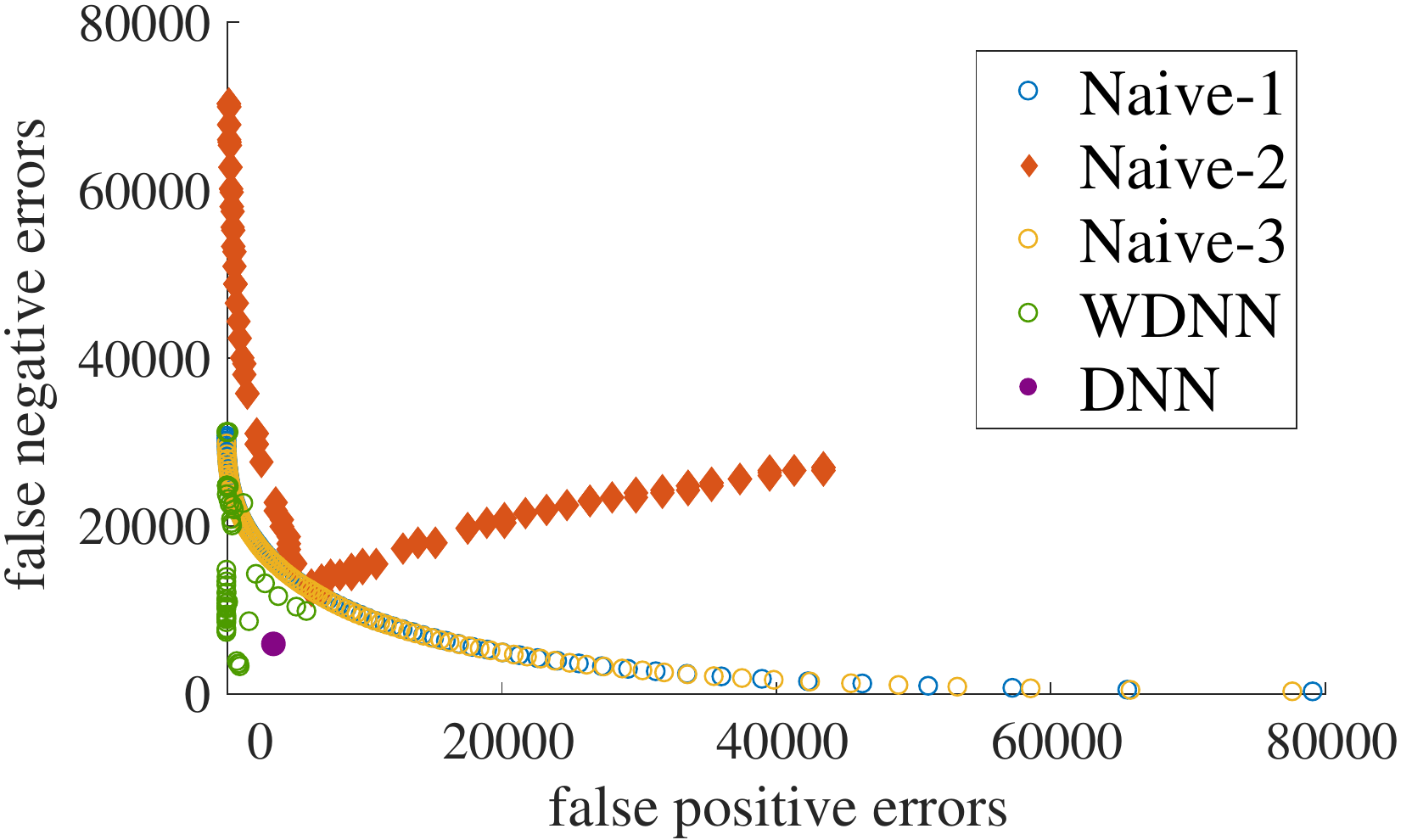}	
	\caption{False positives vs.~false negatives for complex network II}
	\label{fig:prdt_complex-1-2-5-3-results}
\end{figure}

Figure \ref{fig:prdt_complex-1-2-5-3} shows the complex network with three retailers and Figure~\ref{fig:prdt_complex-1-2-5-3-results} 
presents the corresponding results of each algorithm.  

\begin{figure}[H]
	\centering
	\includegraphics[scale=0.4]{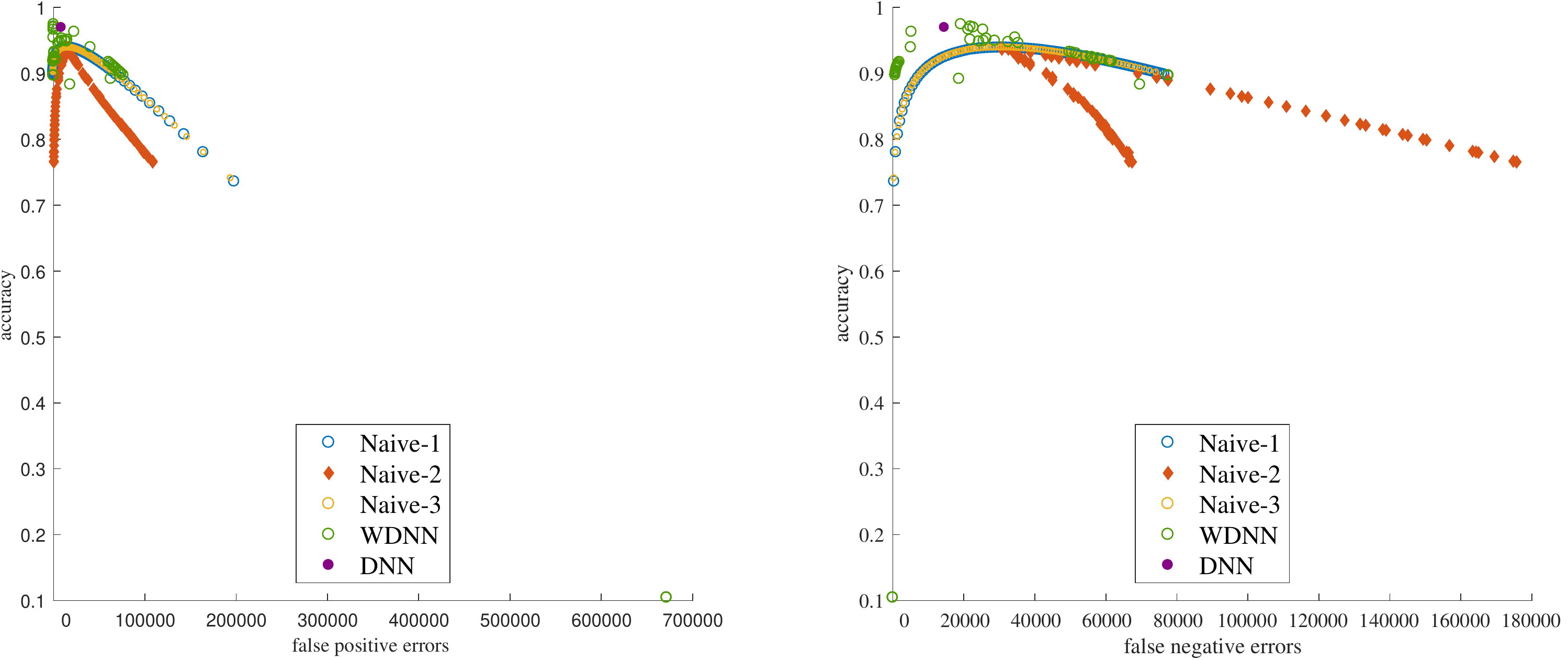}
	\caption{Accuracy of each algorithm for complex network II}
	\label{fig:prdt_complex_1_2-accuracy}
\end{figure}

Figure~\ref{fig:prdt_complex-1-2-5-3-results} plots the false-negative errors vs.~the false-positive errors for each approach and for a range of $\alpha$ values for the naive approaches and a range of weights for the weighted DNN approach. 
Figure~\ref{fig:prdt_complex_1_2-accuracy} plots the errors vs. the accuracy of the predictions.
As we did for the other network topologies, for complex network II we tuned the DNN network hyper-parameters for the case of $c_p = 2$ and $c_n = 1$ and used the resulting hyper-parameters for all other values of $(c_p,c_n)$. 
\add{However, the hyper-parameters obtained in this way did not work well for 46 sets of $(c_p,c_n)$ values, mostly those with $c_p=1$. 
In these cases, the training network did not converge, i.e., after 3 epochs of training, the network generally predicted 0 (or 1) for every data instance, even in the training set, and the loss values failed to decrease to an acceptable level.
Thus, we also tuned the hyper-parameters for $c_p = 1$ and $c_n = 11$ and used them to obtain the results for these 46 cases. 
The hyper-parameters obtained using $(c_p,c_n)=(2,1)$ and $(c_p,c_n)=(1,11)$ are all given in Table~\ref{tb:all_parameters}. We used the first set of hyper-parameters for 72 of the 118 combinations of $(c_p,c_n)$ values and the second set for the remaining 46 combinations. 
Additional hyper-parameter tuning would result in further improved dominance of the DNN approach. 

}

{   
\singlespacing
\begin{table}
	\centering
	\caption{The hyper-parameters used for each network}
	\label{tb:all_parameters}	
	\begin{tabular}{l|ccc}
		Network & lr & $\gamma$ & $\lambda$ \\ \hline
		Serial & 0.001 & 0.0005 & 0.0001 \\		
		Distribution & 0.0005 & 0.001 & 0.0005 \\		
		OWMR & 0.001 & 0.0005 & 0.0005 \\
		Complex-I & 0.05 & 0.000005 & 0.000005 \\
		Complex-II, $(c_p,c_n)=(2,1)$ & 0.05 & 0.05 & 0.05 \\	
		Complex-II, $(c_p,c_n)=(1,11)$ & 0.005 & 0.005 & 0.005 \\	 \hline	
	\end{tabular}
\end{table}
}

Complex network II is the most complex network among all the networks we analyzed, since it is a non-tree network with multiple retailers. As Figure \ref{fig:prdt_complex_1_2-accuracy} shows, WDNN performs worse than the naive approaches for a few values of the weight, which shows the difficulty of the problems and the need to tune the network's hyper-parameters for each set of cost coefficients.

\subsection{Results: Comparison}

In order to get more insight, the average accuracy of each algorithm for each of the networks is presented in Table \ref{prdt:tb:average-multi-echelon}. The average is taken over all instances of a given network type, i.e., over all cost parameters. In the column headers, N1, N2, and N3 stand for the Naive-1, Naive-2, and Naive-3 algorithms. \add{The corresponding hyper-parameters that we used to obtain these results are also presented in Table \ref{tb:all_parameters}.}

DNN provides the best accuracy compared to the other algorithms. WDNN is equally good for the serial and OWMR networks and slightly worse for the distribution and complex II networks. The difference is larger for complex I; this is a result of the fact that we did not re-tune the DNN network for each value of the cost parameters, as discussed in Section~\ref{sec:prdt_section_result_complexI}. We conclude that DNN is the method to choose if the user wants to ensure high accuracy; and WDNN is useful if the user wants to control the balance between false positive and false negative errors.

The column labeled $\text{N3}<\text{N1}$ shows the number of cost-parameter values in which one of Naive-3's predictions has fewer false positive and fewer false negative errors than at least one of the predictions of Naive-1. This happens often for some networks, since the simulated data are normally distributed and since Naive-3 happens to assume a normal distribution. We would expect the method to work worse if the simulated data were from a different distribution. 

The last column shows a similar comparison for the Naive-3 and WDNN algorithms. In particular, Naive-3 never dominates WDNN in this way.

{   
\singlespacing
\begin{table}[]
	\centering
	\caption{Average accuracy of each algorithm}
	\label{prdt:tb:average-multi-echelon}
	\begin{tabular}{l|ccccccc}
		Network & N1 & N2 & N3 & WDNN & DNN  & $\text{N3}<\text{N1}$ & $\text{N3}<\text{WDNN}$\\
		\hline
		Serial       & 0.94    & 0.97    & 0.95    & 0.99 & 0.99 & 0 & 0 \\
		Distribution       & 0.91    & 0.93    & 0.99    & 0.95 & 0.95 & 91 & 0 \\
		OWMR    & 0.91    & 0.93    & 0.91    & 0.95 & 0.98 & 19 & 0 \\		
		Complex I    & 0.86    & 0.94    & 0.92    & 0.91 & 0.97 & 22 & 0  \\ 
		Complex  II   & 0.86    & 0.94    & 0.92    & 0.94 & 0.97 &  0 & 0  \\ \hline
	\end{tabular}
\end{table}
}

\subsection{Extended Results}\label{sec:prdt:extentions}

In this section we present results on some extensions of our original model and analysis. In Section~\ref{sec:prdt:extebded_result_ind10}, we examine the ability of the algorithms to predict whether the inventory level will fall below a given threshold that is not necessarily 0. In Section~\ref{sec:prdt:result_dependent_demand}, we apply our method to problems with dependent demands. Finally, in Section~\ref{sec:prdt:result_multi_period}, we explore multiple-period-ahead predictions.

\subsubsection{Threshold Prediction}\label{sec:prdt:extebded_result_ind10}

The models discussed above aim to predict whether a stock-out will occur; that is, whether the inventory level will fall below 0. However, it is often desirable for inventory managers to have more complete knowledge about inventory levels; in particular, we would like to be able to predict whether the inventory level will fall below a given threshold that is not necessarily 0. In order to see how well our proposed algorithms perform at this task, in this section we provide results for the case in which we aim to predict whether the inventory level will fall below 10. 

\begin{figure}[H]
	\centering
	\includegraphics[scale=0.55]{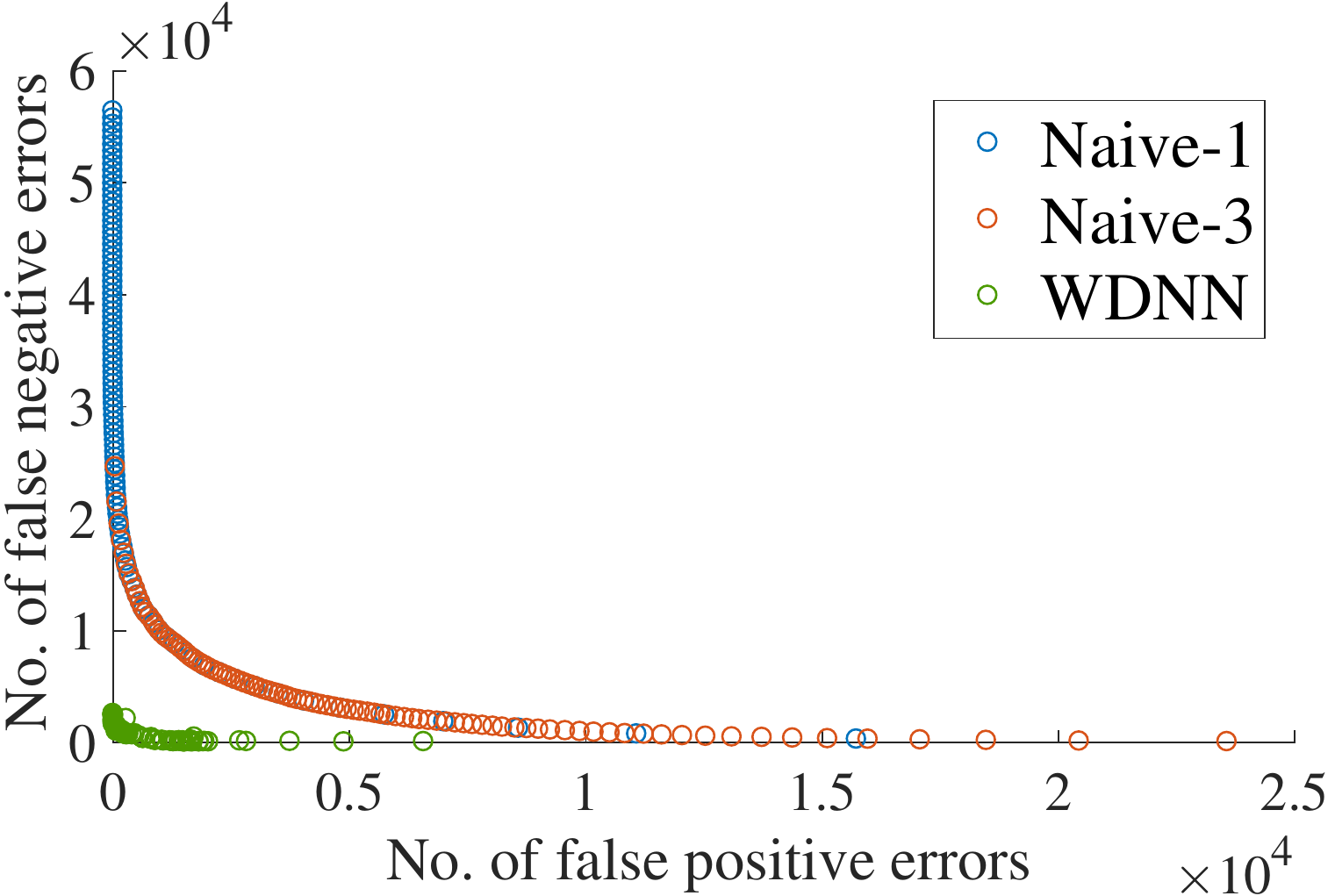}
	\caption{False positives vs.~false negatives for serial network}
	\label{fig:prdt_serial-weighted-results-10}
\end{figure}
\begin{figure}[H]
	\centering
	\includegraphics[scale=0.55]{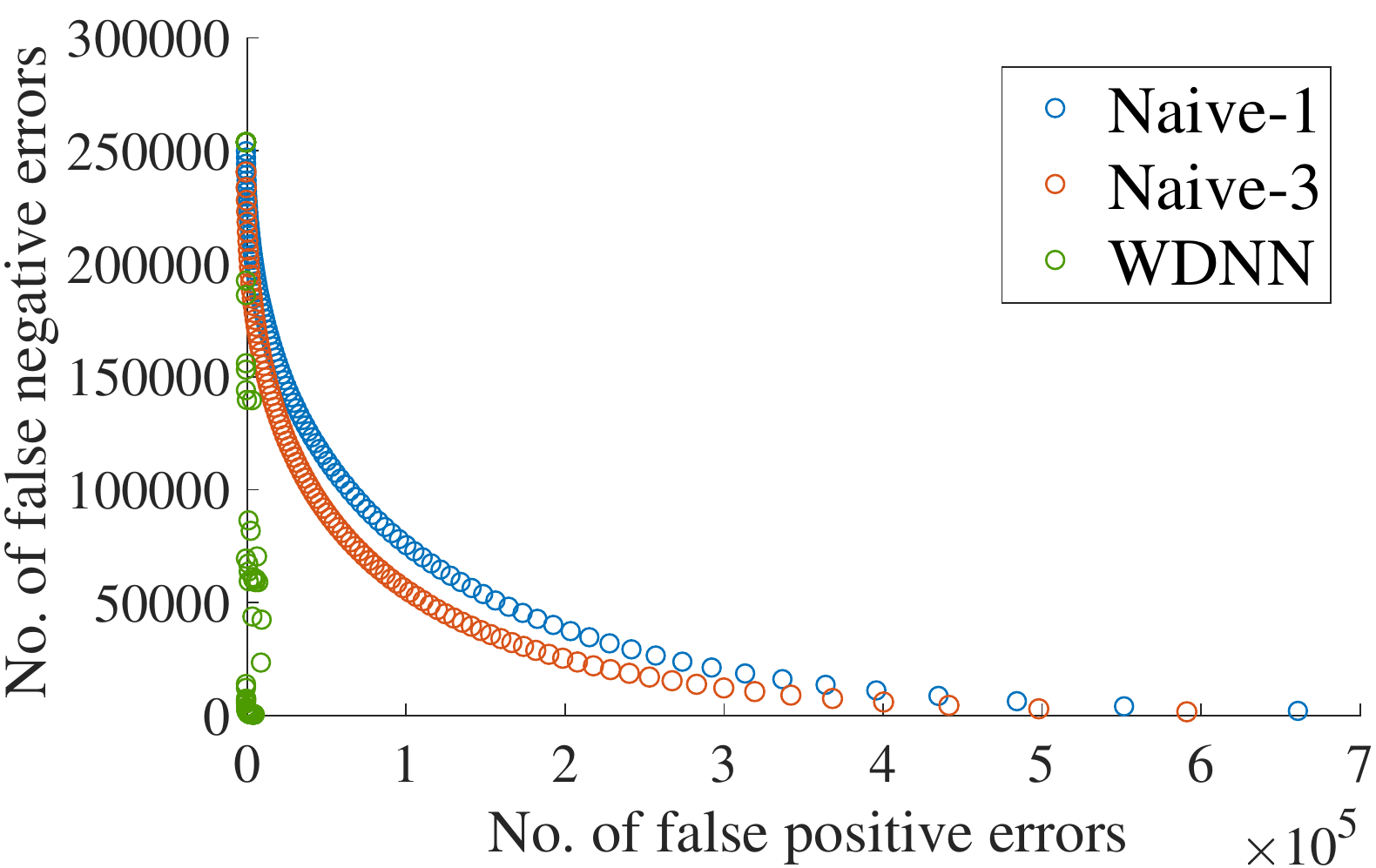}
	\caption{False positives vs.~false negatives for OWMR network}
	\label{fig:prdt_tree-results-10}
\end{figure}
\begin{figure}[H]
	\centering
	\includegraphics[scale=0.55]{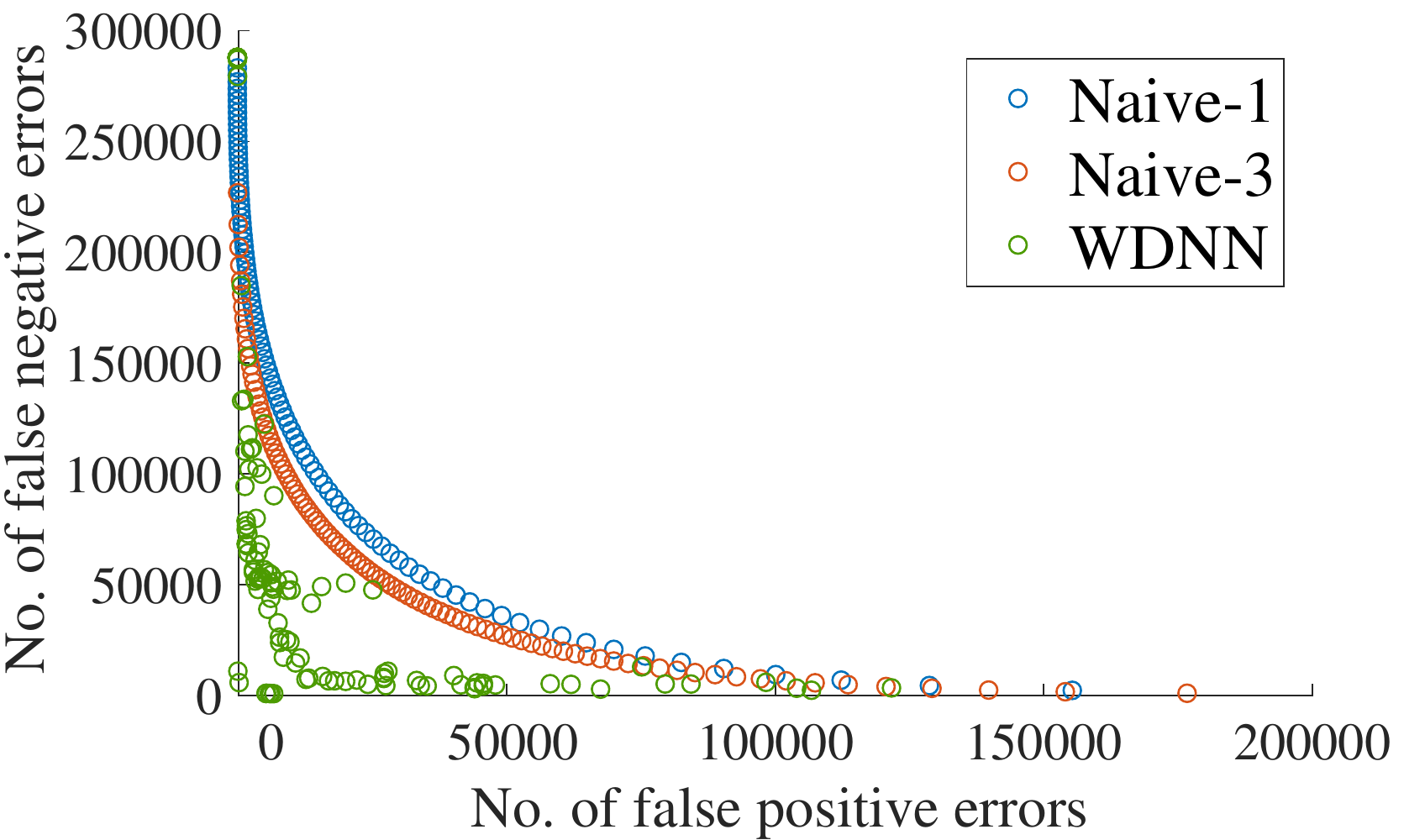}
	\caption{False positives vs.~false negatives for distribution network}
	\label{fig:prdt_distribution_weighted_DNN_Naive-10}
\end{figure}
\begin{figure}[H]
	\centering
	\includegraphics[scale=0.55]{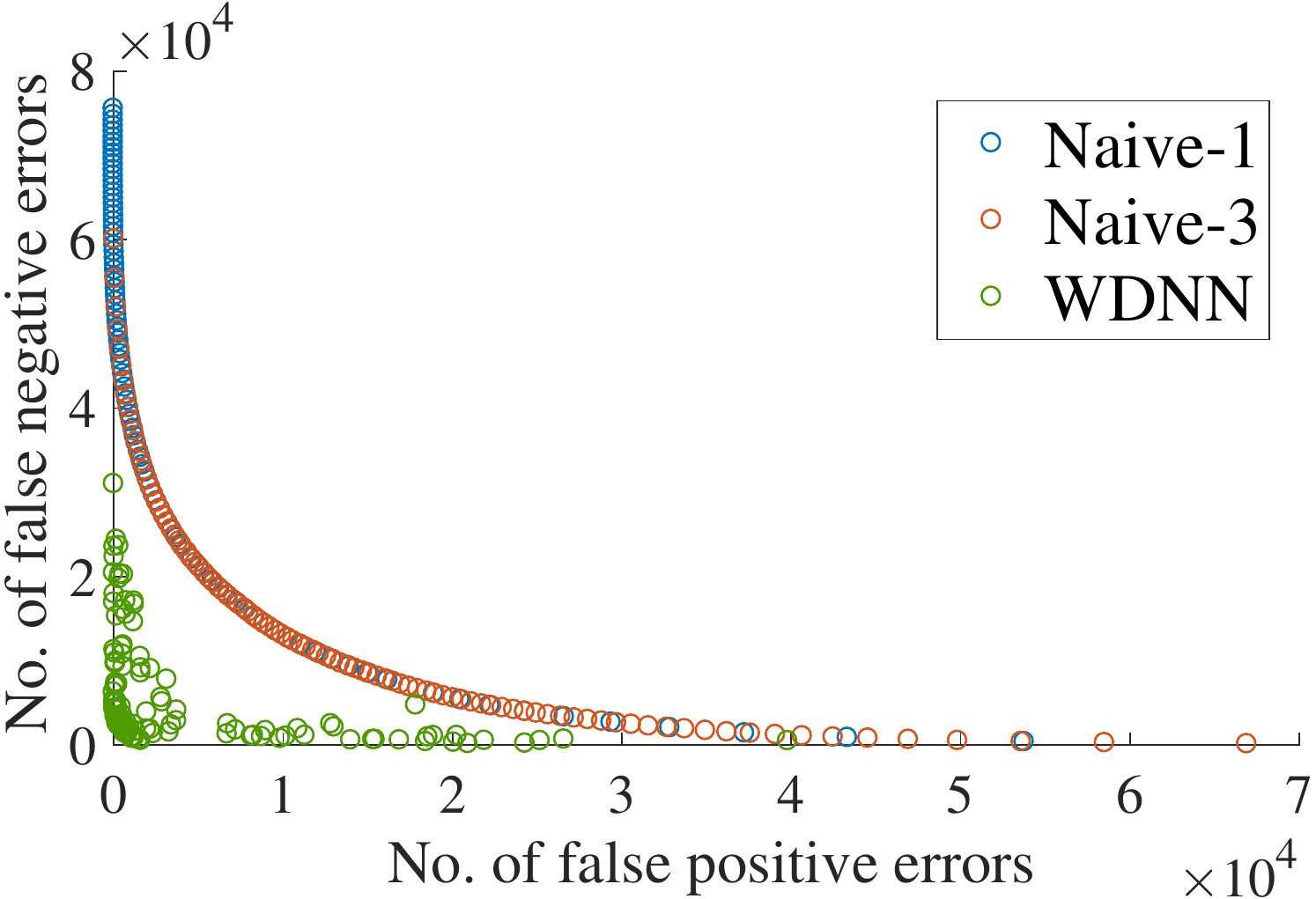}
	\caption{False positives vs.~false negatives for complex network I}
	\label{fig:prdt_complex_2_2-accuracy-10}
\end{figure}
\begin{figure}[H]
	\centering
	\includegraphics[scale=0.55]{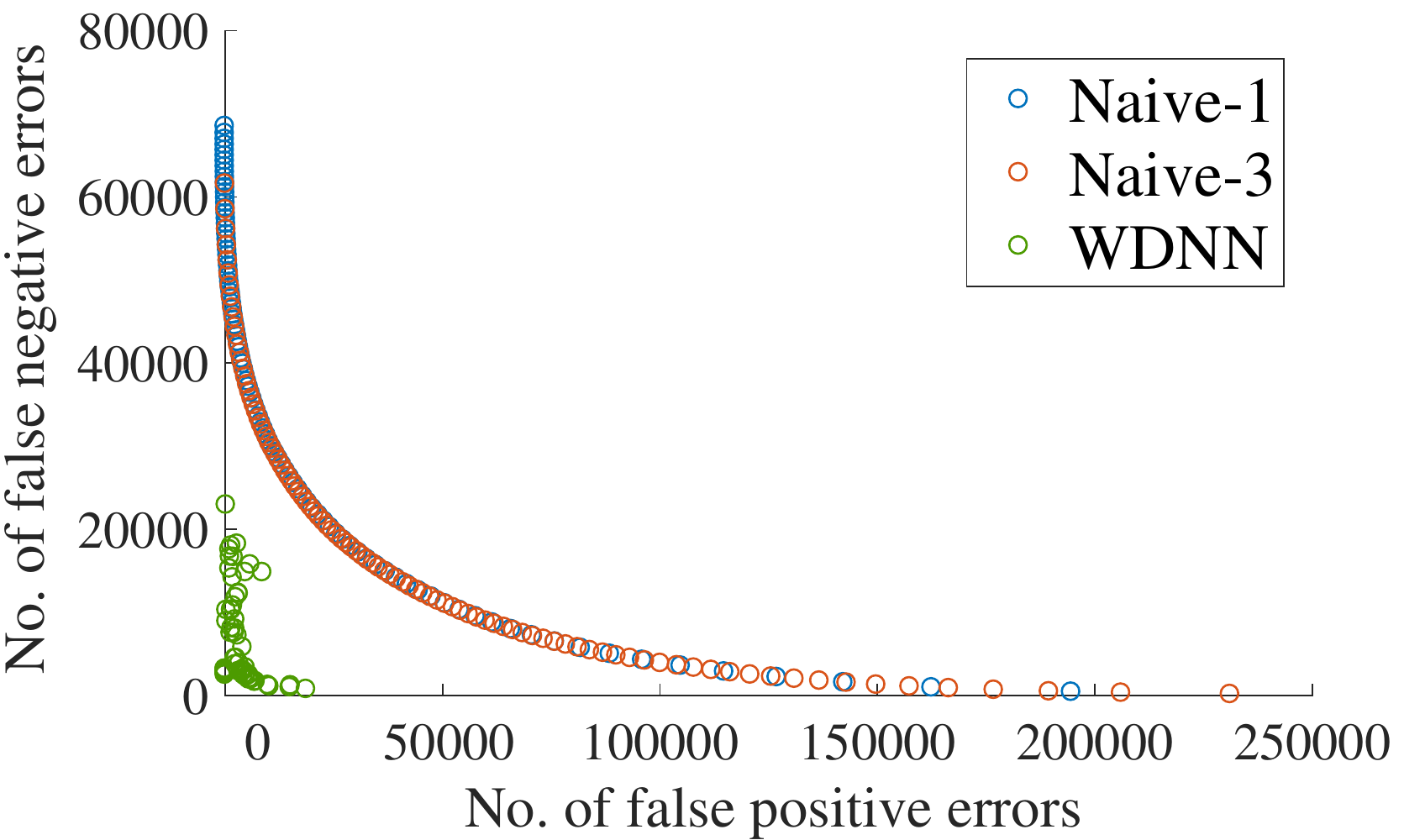}	
	\caption{False positives vs.~false negatives for complex network II}
	\label{fig:prdt_complex-1-2-5-3-results-10}
\end{figure}

A similar procedure 
is applied to achieve the results of all algorithms. In particular, we changed the way that the data labels are applied so that we assign a label of 1 when $IL<10$ and a label of 0 otherwise. We exclude the results of the DNN and Naive-2 algorithms, since they are dominated by the  WDNN and Naive-3 algorithms. 
Figures \ref{fig:prdt_serial-weighted-results-10}--\ref{fig:prdt_complex-1-2-5-3-results-10} 
present the results of the serial, OWMR, distribution, complex I, and complex II networks. 
As before, WDNN outperforms the naive algorithms. 
Table \ref{prdt:tb:average-multi-echelon-ind10} provides the overall accuracy of all algorithms and the comparisons between them; the columns are the same as those in Table~\ref{prdt:tb:average-multi-echelon}. As before, WDNN performs better than or equal to the other algorithms for all networks. The  accuracy figures for this case are provided in Appendix \ref{sec:prdt:appdx:results_ind10}.

{   
\singlespacing
\begin{table}[]
	\centering
	\caption{Average accuracy of each algorithm for predicting inventory level less than 10}
	\label{prdt:tb:average-multi-echelon-ind10}
	\begin{tabular}{l|ccccc}
		Network & N1 & N3 & WDNN & $\text{N3}<\text{N1}$ & $\text{N3} < \text{WDNN}$\\
		\hline
		Serial       & 0.88     & 0.96    & 0.99 & 0  & 0 \\
		Distribution  & 0.90    & 0.92    & 0.93 & 89 & 0\\
		OWMR    	 & 0.91     & 0.92    & 0.96 & 97 & 0 \\
		Complex I    & 0.85    & 0.87    & 0.97 & 13 & 0  \\ 
		Complex  II   & 0.82     & 0.87    & 0.96 &  0 & 0  \\
		\hline
	\end{tabular}
\end{table}
}

\subsubsection{Multi-Item Dependent Demand Multi-Echelon Problem}\label{sec:prdt:result_dependent_demand} 

The data sets we have used so far assume that the demands are statistically independent. However, in the real world, demand for multiple items are often dependent on each other. 
Moreover, this dependence information \add{provides additional information for DNN and} might help to provide more accurate stock-out predictions. To analyze this, we generated the data for seven items with dependent demands, some positively and some negatively correlated.
\add{The mean demand of the seven items for seven days of a week is shown in Figure~\ref{fig:prdt_distributions_demand_weekly}.
For more details see Appendix~\ref{sec:prdt:appdx:dependent_data_generation} which provides the demand means and standard deviation for each item and each day.}

	
	\begin{figure}[H]
		\centering
		\includegraphics[scale=0.3]{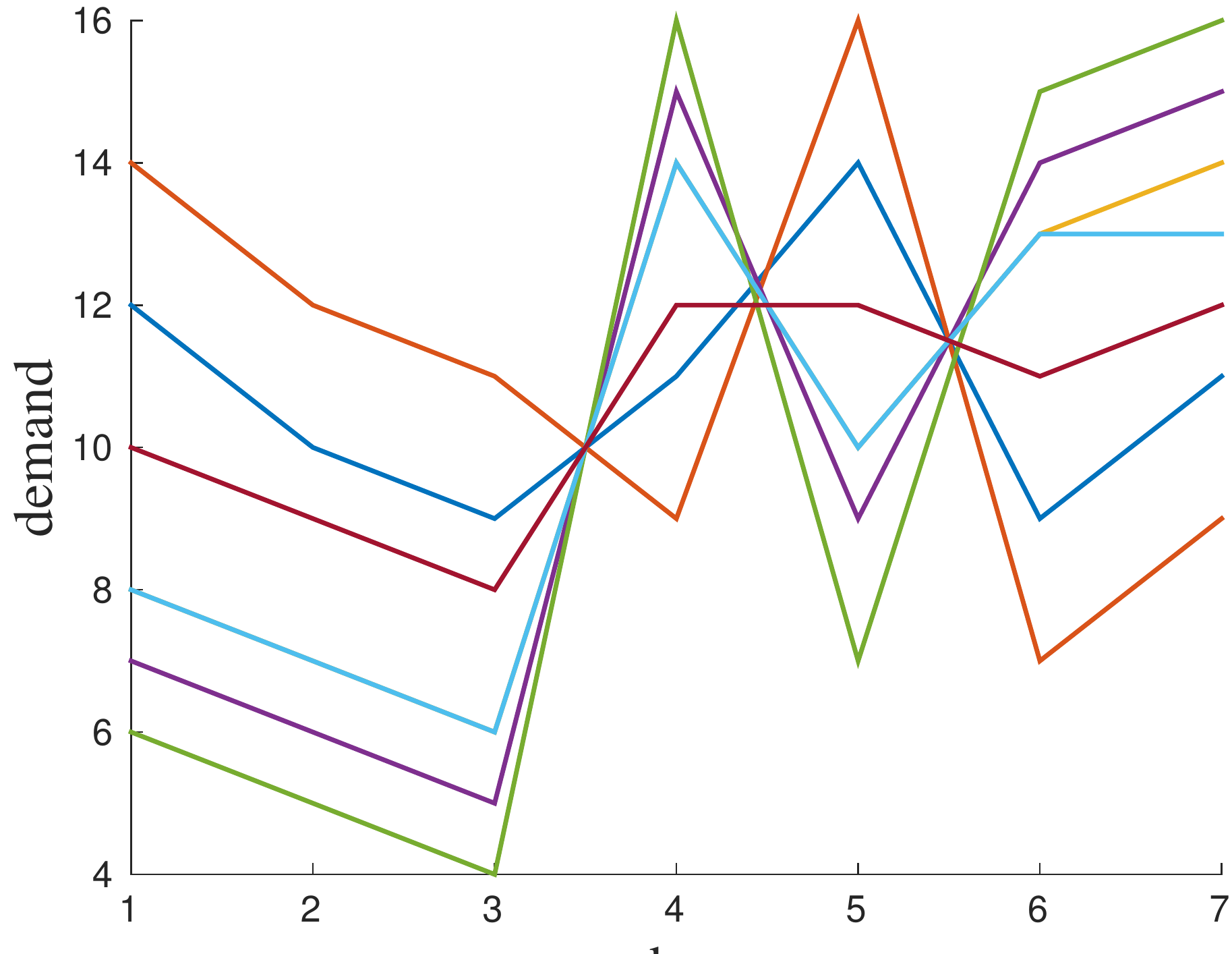}
		\caption{The demand of seven items in each day}
		\label{fig:prdt_distributions_demand_weekly}
	\end{figure}

\begin{figure}[H]
	\centering
	\includegraphics[scale=0.55]{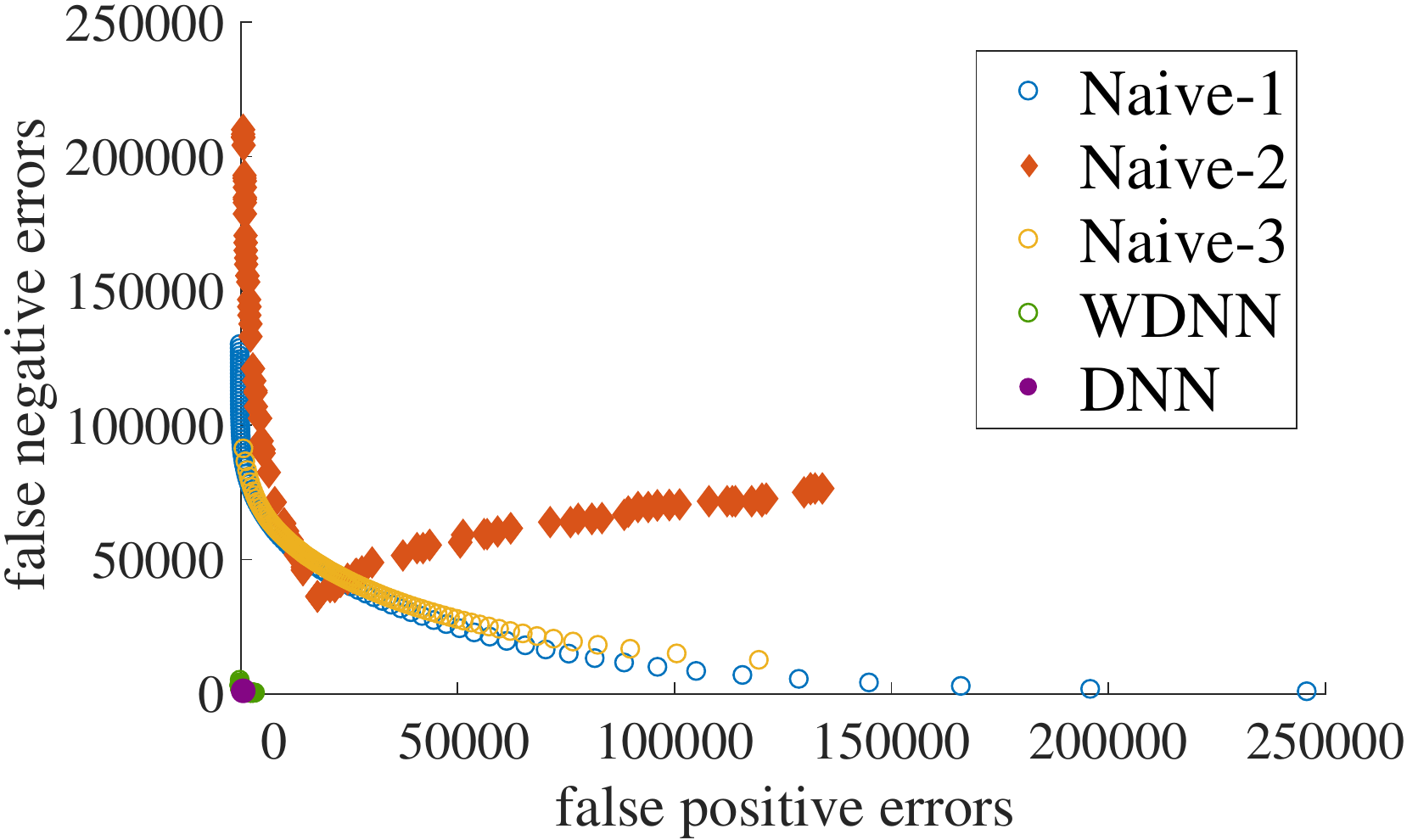}
	\caption{False positives vs.~false negatives for distribution network with multi-item dependent demand}
	\label{fig:prdt_distribution_weekly_weighted_DNN_Naive}
\end{figure}

\begin{figure}[H]
	\centering
	\includegraphics[scale=0.4]{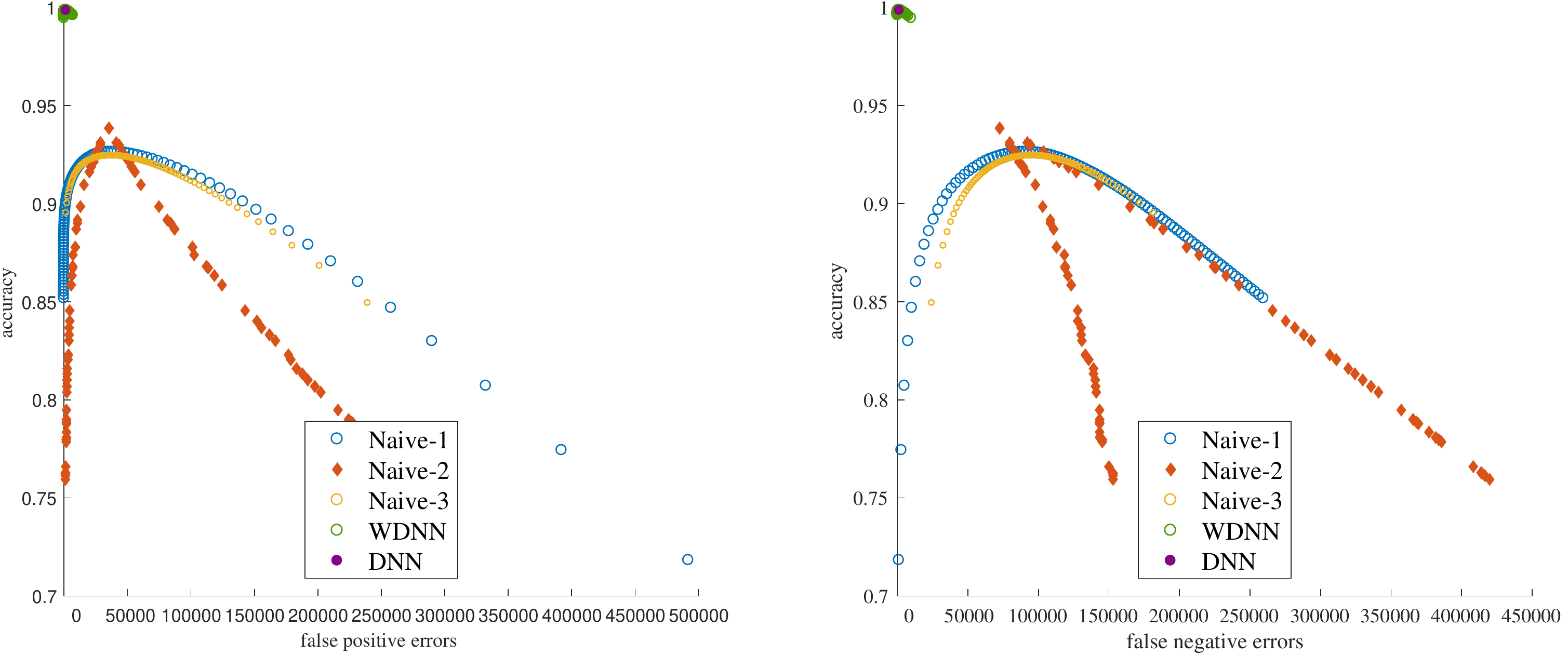}
	\caption{Accuracy of each algorithm for distribution network with multi-item dependent demand}
	\label{fig:prdt_distribution-weekly-accuracy}
\end{figure}

\add{ We tested this approach using the distribution network (Figure~\ref{fig:prdt_distribution-1-2-3-7})}.
Figure~\ref{fig:prdt_distribution_weekly_weighted_DNN_Naive} plots the false-negative errors vs.~the false-positive errors for each approach and for a range of $\alpha$ values for the naive approaches and a range of weights for the weighted DNN approach. 
WDNN produces an average accuracy rate of 99\% for this system, compared to 95\% for the independent-demand case, which shows how DNN is able to make more accurate predictions by taking advantage of information it learns about the demand dependence.
Finally, Figure~\ref{fig:prdt_distribution-weekly-accuracy} plots the errors vs. the accuracy of the predictions. DNN and WDNN provide much more accurate predictions than the naive methods.

\subsubsection{Multi-Period Prediction} \label{sec:prdt:result_multi_period}

In order to see how well our algorithm can make stock-out predictions multiple periods ahead, we revised the DNN structure, such that there are $n \times q$ output values in the DNN algorithm, where $q$ is the number of prediction periods. We tested this approach using  the distribution network (Figure \ref{fig:prdt_distribution-1-2-3-7}). 

\begin{figure}[H]
	\centering
	\includegraphics[scale=0.5]{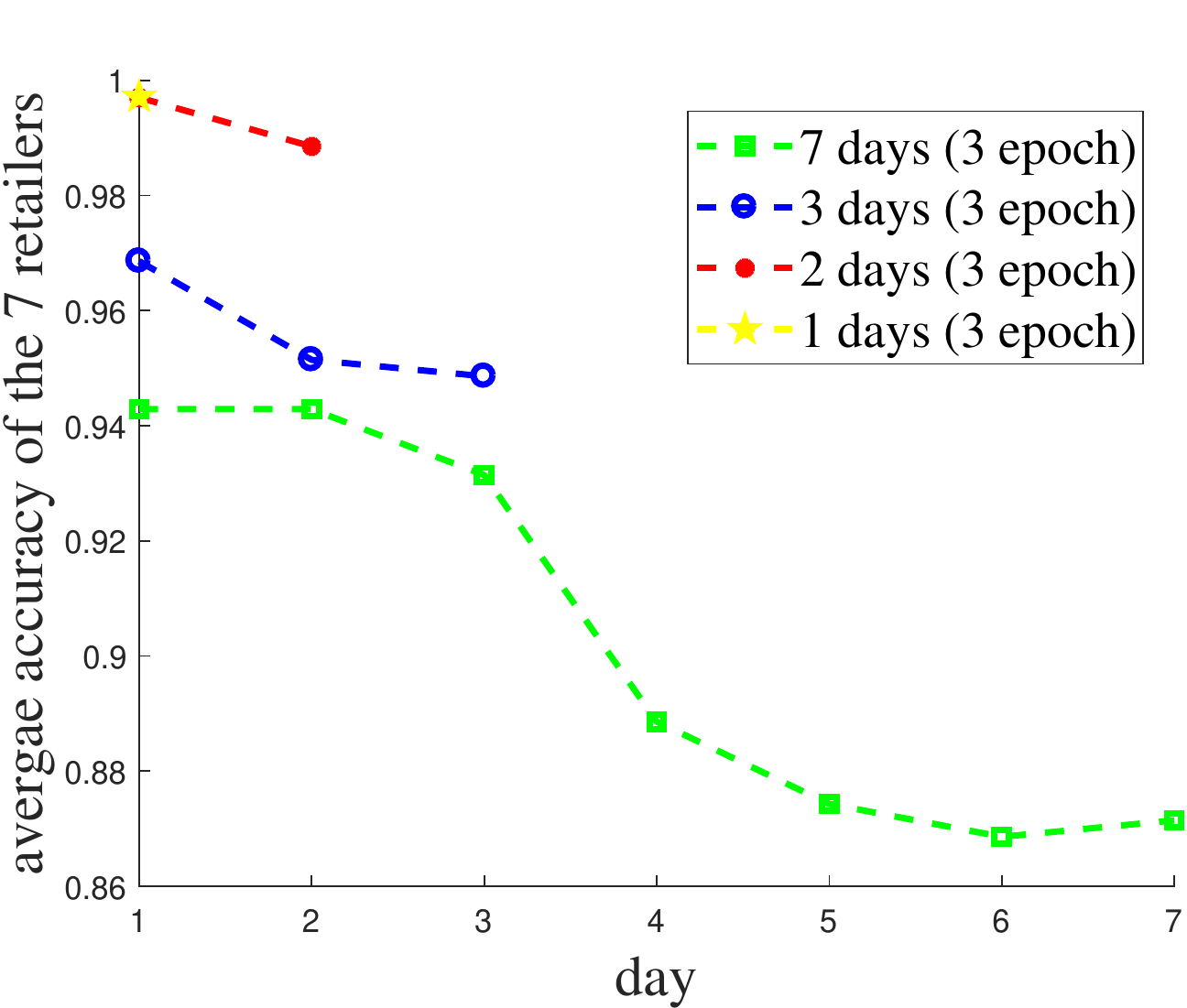}
	\caption{Average accuracy over seven days in multi-period prediction}
	\label{fig:distributions_multiple_period_retailers_avg}
\end{figure}

We tested the algorithm for three different problems. The first predicts stock-outs for each of the next two days; the second and third to the same for the next three and seven days, respectively. 
The accuracy of the predictions for each day are plotted in Figure \ref{fig:distributions_multiple_period_retailers_avg}. For example, the blue curve shows the accuracy of the predictions made for each of the next 3 days when we make predictions over a horizon of 3 days. 
The  one-day prediction accuracy is plotted as a reference.

Not surprisingly, it is harder to predict stock-outs multiple days in advance. For example, the accuracy for days 4--7 is below 90\% when predicting 7 days ahead. Moreover, when predicting over a longer horizon, the predictions for earlier days are less accurate. For example, the accuracy for predictions 2 days ahead is roughly 99\% if we use a 2-day horizon, 95\% if we use a 3-day horizon, and 94\% if we use a 7-day horizon. Therefore, if we wish to make predictions for each of the next $q$ days, it is more accurate (though slower) to run $q$ separate DNN models rather than a single model that predicts the next $q$ days.

\section{Conclusion and Future Works}\label{sec:prdt_conclusions}

We studied stock-out prediction in multi-echelon supply chain networks. In single-node networks, classical inventory theory provides tools for making such predictions when the demand distribution is known.  However, there is no algorithm to predict stock-out in multi-echelon networks. 
To address this need, we proposed an algorithm based on deep learning. We also introduced three naive algorithms  to provide a benchmark for stock-out prediction. 
None of the algorithms require knowledge of the demand distribution; they use only historical data. 

Extensive numerical experiments show that the DNN algorithm works well compared to the three naive algorithms. 
The results suggest that our method holds significant promise for predicting stock-outs in complex, multi-echelon supply chains. It obtains an average accuracy of 99\% in serial networks and 95\% for OWMR and distribution networks. Even for complex, non-tree networks, it attains an average accuracy of at least 91\%. It also performs well when predicting inventory levels below a given threshold (not necessarily 0), making predictions when the demand is correlated, and making predictions multiple period ahead.

Several research directions are now evident, including expanding the current approach to handle other types of uncertainty, e.g., lead times, supply disruptions, etc. 
Improving the model's ability to make accurate predictions for more than one period ahead is another interesting research direction. Our current model appears to be able to make predictions accurately up to roughly 3 periods ahead, but its accuracy degrades quickly after that. Finally, the model can be extended to take into account other supply chain state variables in addition to current inventory and in-transit levels.

\section{Acknowledgment}
This research was supported in part by NSF grant \#CMMI-1663256. This support is gratefully acknowledged.

\bibliographystyle{plainnat}
\bibliography{multi_echelon_v2}
	

\clearpage

	\appendix
	
\section{Stock-Out Prediction for Single-Stage Supply Chain Network}\label{sec:prdt_appd_one_agent_prediction}

Consider a single-stage supply chain network.
The goal is to obtain the stock-out probability and as a result make a stock-out prediction, i.e., we want to obtain the probability:
\begin{equation*}
P(IL_t < 0),
\end{equation*}
where $IL_t$ is the ending inventory level in period $t$. Classical inventory theory (see, e.g., \cite{snyder2018fundamentals,zipkin2000foundations}) tells us that
\begin{equation*} IL_t = IP_{t-L} - D_L, \end{equation*}
where $L$ is the lead time, $IP_{t-L}$ is the inventory position (inventory level plus on-order inventory) after placing a replenishment order in period $t-L$, and $D_L$ is the lead-time demand. Since we know $IP_{t-L}$ and we know the probability distribution of $D_L$, we can determine the probability distribution of $IL_t$ and use this to calculate $P(IL_t<0)$. Then we can predict a stock-out if this probability is larger than $\alpha$, for some desired threshold $\alpha$.

%

\section{Gradient of Weighted Soft-max Function}\label{sec:prdt_appd_grdt_weighted_softmax}

	Let
	$$p_j = \frac{e^{z_j}}{\sum_{u=1}^U e^{z_u}}.$$
	Then the gradient of the soft-max loss function \eqref{eq:softmax_loss} is:
	\begin{equation*} \frac{\partial E}{\partial z_{j}} = p_j - y_j \end{equation*}
	
	and the gradient of weighted soft-max loss function \eqref{prdt:eq:softmax_loss_weighted} is:
	\begin{equation*} \frac{\partial E_w}{\partial z_{j}} = w_j (p_j - y_j). \end{equation*}

\section{Activation and Loss Functions }\label{sec:prdt:appdx:losses}

The most common loss functions are the hinge \eqref{hinge_loss}, logistic \eqref{logistic_loss}, and Euclidean  \eqref{euclidean_loss} loss functions, given (respectively) by:
\begin{equation}
\label{hinge_loss}
E = \max(0, 1 - y_i \hat{y}_i)
\end{equation}

\begin{equation}
\label{logistic_loss}
E = − \log (1 + e^{−y_i \hat{y}_i })
\end{equation}

\begin{equation}
\label{euclidean_loss}
E = ||y_i - \hat{y}_i ||_2^2,
\end{equation}
where $y_i$ is the observed value of sample $i$, and $\hat{y_i}$ is the output of the DNN.
The hinge loss function is appropriate for $0, 1$ classification. 
The logistic loss function is also used for $0,1$ classification; however, it is a convex function which is easier to optimize than the hinge function.
The Euclidean loss function minimizes the difference between the observed and calculated values and penalizes closer predictions much less than farther predictions.

Each node of the DNN network has an activation function. The most commonly used activation functions are sigmoid, tanh, and inner product, given (respectively) by:
	\begin{align}
 	\text{Sigmoid}(z) & = \frac{1}{1+e^{1+z}} \\
 	\text{Tanh}(z) & = \frac{2e^z - 1}{2e^z+1} \\
 	\text{InnerProduct}(z) & = z
	\end{align}

\section{Dependent Demand Data Generation}\label{sec:prdt:appdx:dependent_data_generation}
This section provides the details of data generation for dependent demands. 
In the case of dependent demand, there are seven items, and the demand mean of each item is different on different days of the week. 
Tables \ref{tb:prdt:dependent_demand_mu} and \ref{tb:prdt:dependent_demand_sigma} provide the mean ($\mu$) and standard deviation ($\sigma$) of the normal distribution of for each item in each day of week. 

\begin{table}
	\centering
	\caption{The mean demand ($\mu$) of each item on each day of the week.}
	\label{tb:prdt:dependent_demand_mu}
	\begin{tabular}{l|lllllll}
		Item & Mon & Tue & Wen & Thu & Fri & Sat & Sun \\ \hline
		1 & 12 & 10 & 9 & 11 & 14 & 9 & 11  \\
		2 & 14 & 12 & 11 & 9 & 16 & 7 & 9 \\
		3 & 8 & 7 & 6 & 14 & 10 & 13 & 14 \\
		4 & 7 & 6 & 5 & 15 & 9 & 14 & 15 \\
		5 & 6 & 5 & 4 & 16 & 7 & 15 & 16 \\
		6 & 8 & 7 & 6 & 14 & 10 & 13 & 13 \\
		7 & 10 & 9 & 8 & 12 & 12 & 11 & 12 \\ \hline
	\end{tabular}
\end{table}

\begin{table}
	\centering
	\caption{The mean standard deviation ($\sigma$) of each item on each day of the week.}
	\label{tb:prdt:dependent_demand_sigma}	
	\begin{tabular}{l|lllllll}
		Item & Mon & Tue & Wen & Thu & Fri & Sat & Sun \\ \hline
		1 & 3 & 2 & 4 & 1 & 2 & 3 & 2 \\
		2 & 4 & 3 & 4 & 1 & 3 & 2 & 1 \\
		3 & 1 & 1 & 2 & 2 & 2 & 4 & 3 \\
		4 & 1 & 1 & 1 & 3 & 1 & 4 & 3 \\
		5 & 1 & 1 & 1 & 2 & 1 & 3 & 3 \\
		6 & 2 & 1 & 1 & 3 & 1 & 3 & 3 \\
		7 & 3 & 2 & 4 & 1 & 2 & 3 & 2 \\ \hline
	\end{tabular}
\end{table}
  
\section{Results of Threshold-Prediction Case}\label{sec:prdt:appdx:results_ind10}
\begin{figure}[H]
	\centering
	\includegraphics[scale=0.4]{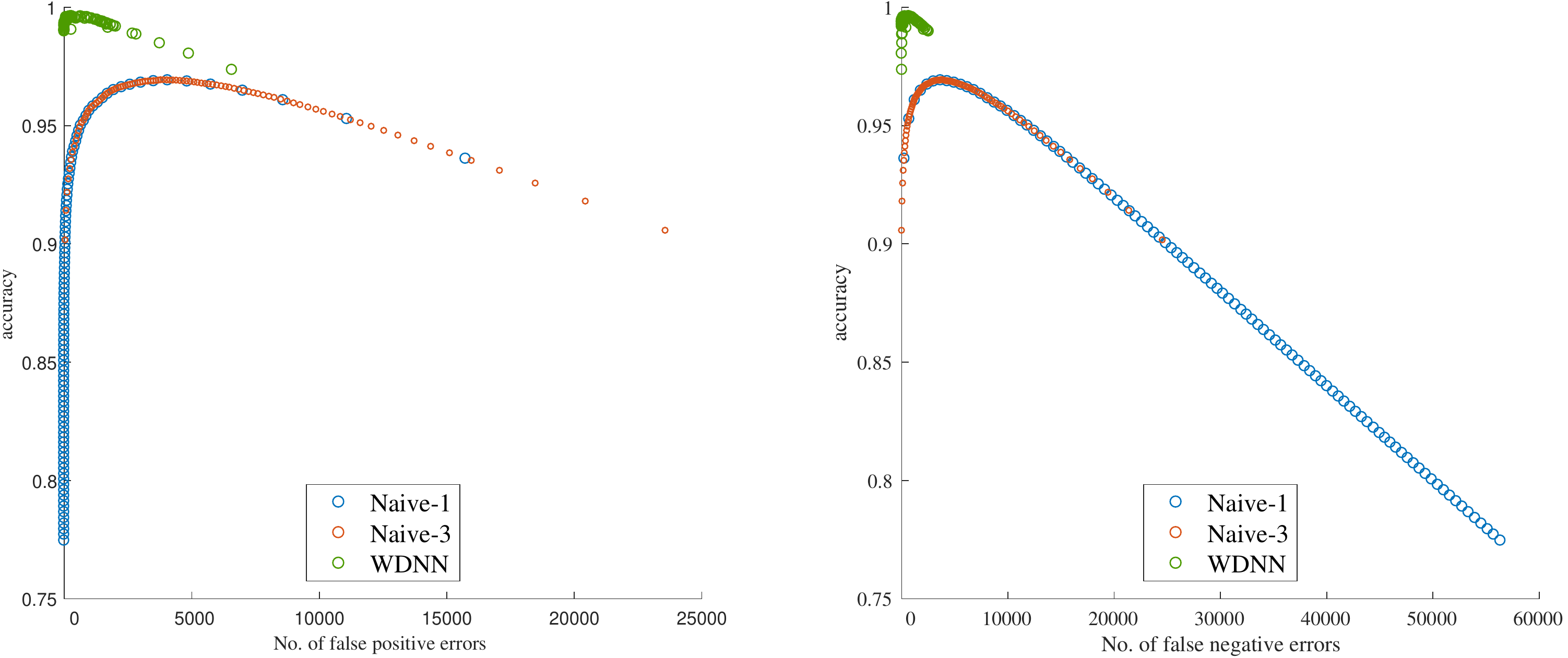}
	\caption{Accuracy of each algorithm for serial network}
	\label{fig:prdt_serial-accuracy-10}
\end{figure}

\begin{figure}[H]
	\centering
	\includegraphics[scale=0.4]{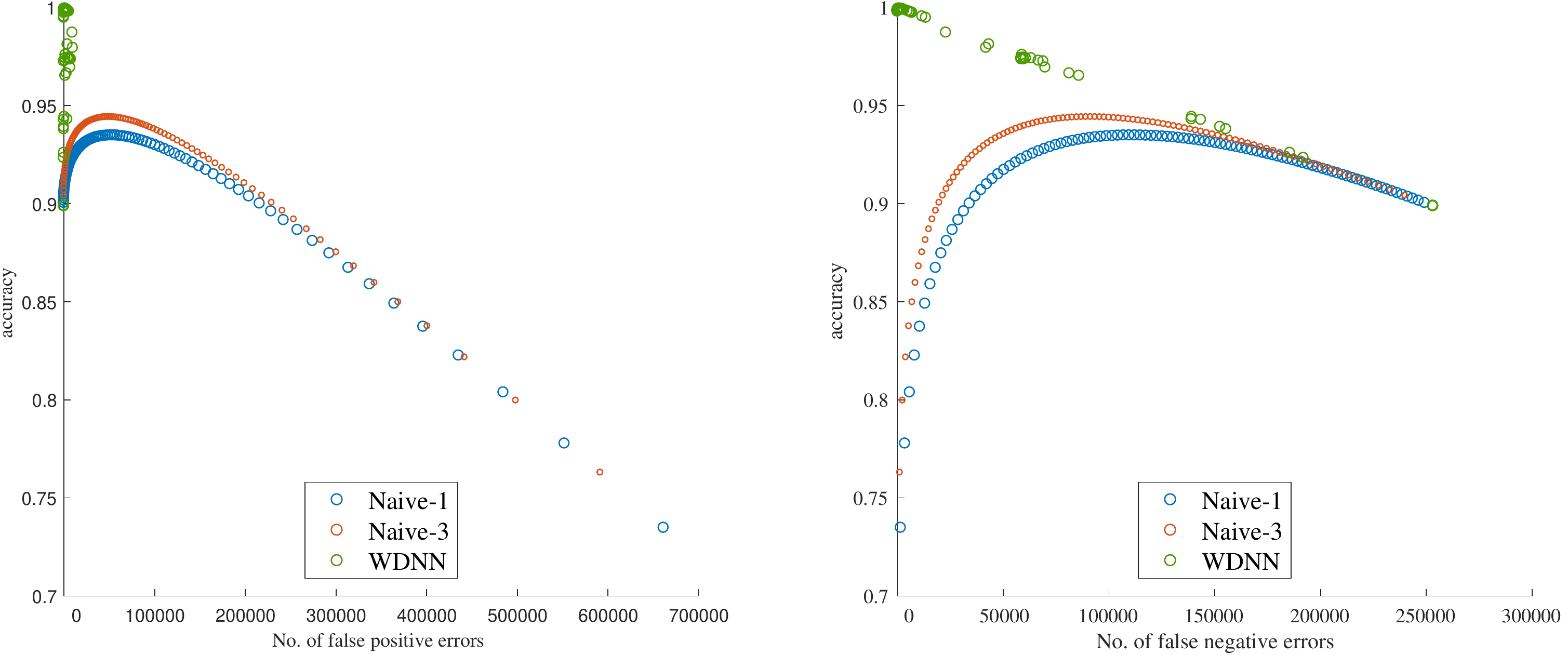}
	\caption{Accuracy of each algorithm for OWMR network}
	\label{fig:prdt_tree-accuracy-10}
\end{figure}

\begin{figure}[H]
	\centering
	\includegraphics[scale=0.4]{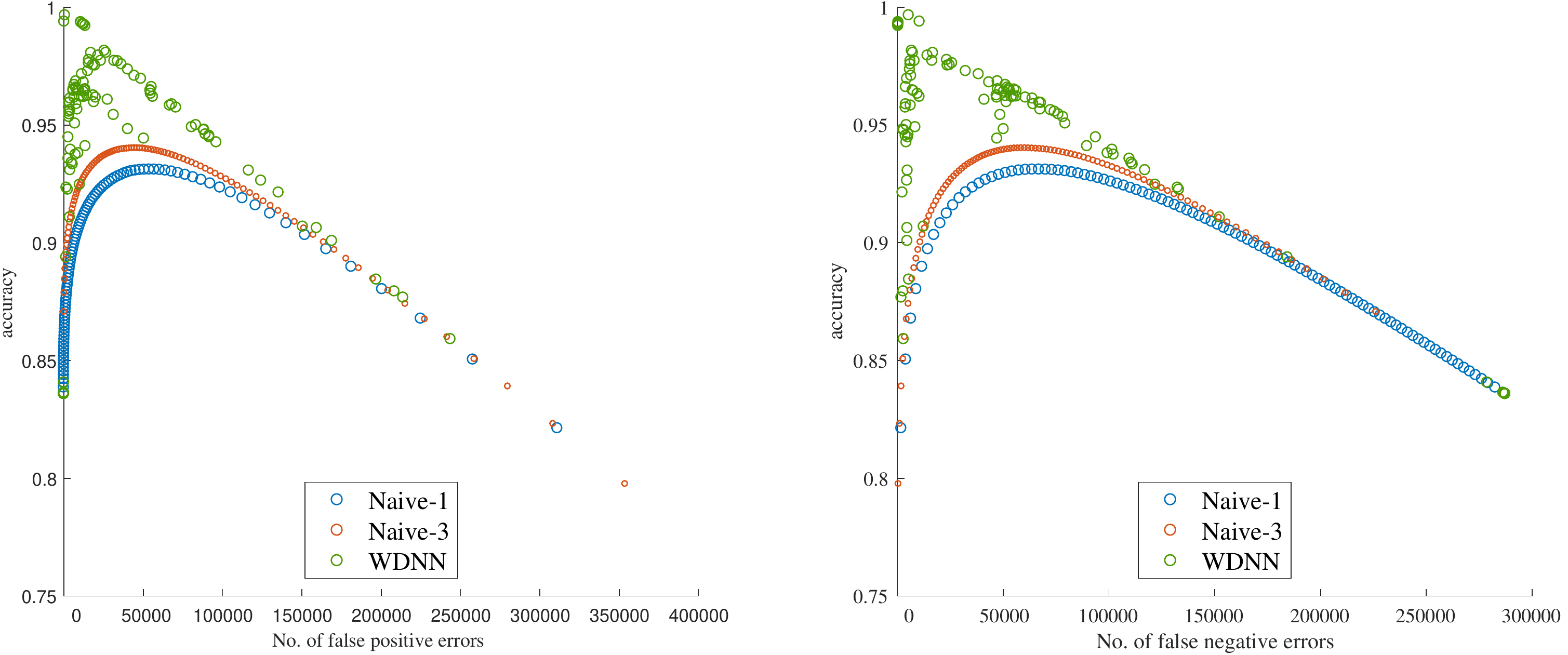}
	\caption{Accuracy of each algorithm for distribution network}
	\label{fig:prdt_distribution-accuracy-10}
\end{figure}

\begin{figure}[H]
	\centering
	\includegraphics[scale=0.4]{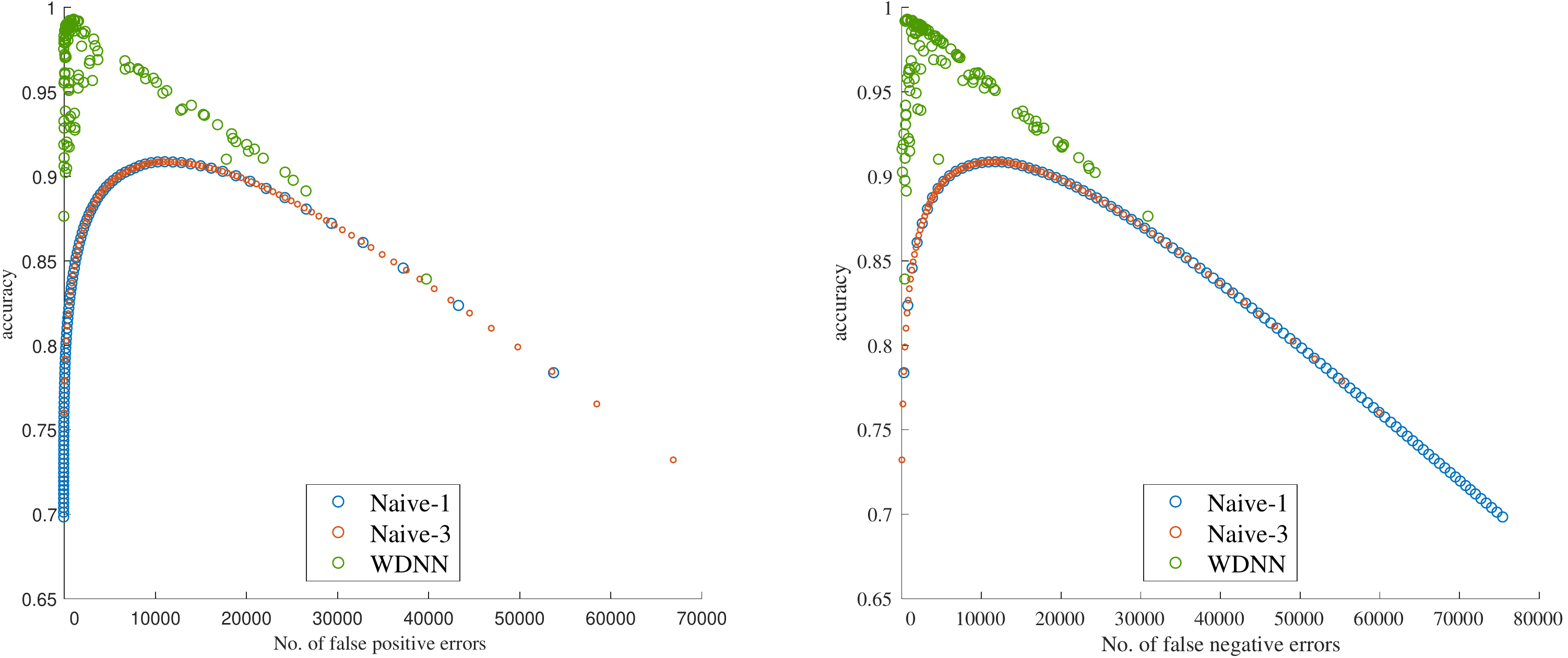}
	\caption{Accuracy of each algorithm for complex network I}
	\label{fig:prdt_complex-2-accuracy-10}
\end{figure}

\begin{figure}[H]
	\centering
	\includegraphics[scale=0.4]{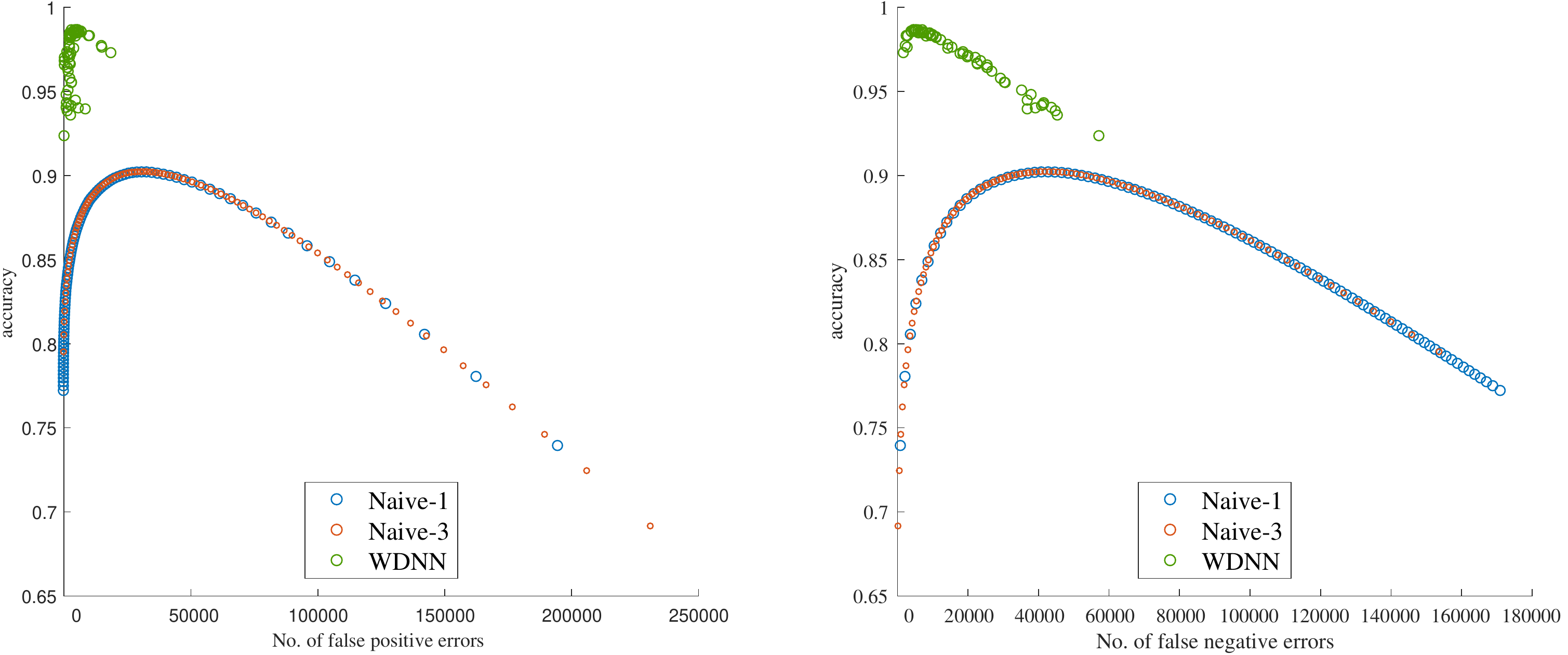}
	\caption{Accuracy of each algorithm for complex network II}
	\label{fig:prdt_complex_1_2-accuracy-10}
\end{figure}

This section provides the accuracy results  for the problem described in Section~\ref{sec:prdt:extebded_result_ind10}, in which we wish to predict whether the inventory level will fall below 10. Figures \ref{fig:prdt_serial-accuracy-10}--
\ref{fig:prdt_complex_1_2-accuracy-10} show the results for the serial, OWMR, distribution, complex I, and complex II networks, respectively.

\end{document}